\documentclass{article} 
\usepackage[preprint]{colm2025_conference}
\usepackage{microtype}
\usepackage{hyperref}
\usepackage{url}
\usepackage{booktabs}

\usepackage{lineno}

\definecolor{darkblue}{rgb}{0, 0, 0.5}
\hypersetup{colorlinks=true, citecolor=darkblue, linkcolor=darkblue, urlcolor=darkblue}

\usepackage[utf8]{inputenc} 
\usepackage[T1]{fontenc}    
\usepackage{wrapfig2}
\usepackage{amsfonts}       
\usepackage{amsmath}        
\usepackage{float}
\usepackage{graphicx}
\usepackage{nicefrac}       
\usepackage{xcolor}         
\usepackage{pgfplots}       
\usepackage{pgfplotstable}  
\usepackage{enumitem}
\usepackage{amssymb}
\usepackage{listings}       
\usepackage{pythonhighlight}
\usepackage{mathtools}
\usepgfplotslibrary{groupplots}

\usepackage{amsmath,amsfonts,bm}

\def\Figref#1{Figure~\ref{#1}}

\def\eqref#1{equation~\ref{#1}}

\def\1{\bm{1}}

\DeclareMathAlphabet{\mathsfit}{\encodingdefault}{\sfdefault}{m}{sl}
\SetMathAlphabet{\mathsfit}{bold}{\encodingdefault}{\sfdefault}{bx}{n}

\newcommand{\softmax}{\mathrm{softmax}}

\title{CE-U: Cross Entropy Unlearning}

\author{Bo Yang \\ Tacnode US Inc \\ \texttt{bo@tacnode.io}}

\begin{document}

\ifcolmsubmission
\linenumbers
\fi

\maketitle

\begin{abstract}
  Large language models memorize sensitive data from their pretraining corpora \cite{jang2022knowledge}. In this work, we propose CE-U (Cross Entropy Unlearning), a loss function for unlearning. CE-U addresses fundamental limitations of gradient ascent approaches that suffer from vanishing gradients when model confidence is high and exploding gradients when confidence is low. We also unify standard cross entropy learning and unlearning into a single framework. On the TOFU benchmark for unlearning \cite{maini2024tofu}, CE-U achieves state-of-the-art results on LLaMA2-7B models without using an extra oracle model or additional positive samples. Our analysis reveals that the problematic gradient ascent component also exists in reinforcement learning algorithms like DPO \cite{rafailov2023direct} and GRPO \cite{Shao2024DeepSeekMath}. This suggests that applying CE-U approach to reinforcement learning could be promising to improve stability and convergence.
\end{abstract}

\section{Introduction}

Large language models memorize sensitive data \cite{jang2022knowledge}. Privacy regulations such as GDPR emphasize the need for data unlearning \cite{Voigt2017}. With pressures like the Right to be Forgotten and related ethical concerns, methods for unlearning data without full retraining have become crucial \cite{Cao2015Forgetting}. Prior work on unlearning has relied on gradient ascent techniques that increase the loss on memorized data \cite{bourtoule2021machine}. However, such approaches suffer from instability and require explicit or implicit regularization, such as additional positive samples or KL penalties against an oracle model.

In this paper, we introduce CE-U (Cross Entropy Unlearning), a novel loss function that leverages a modified cross entropy formulation. By setting the logit corresponding to the true label to negative infinity, CE-U effectively suppresses the model’s confidence in that label while distilling the desired output distribution through KL divergence. Our key contributions are summarized as follows:

\begin{enumerate}[leftmargin=*]
  \item \textbf{Novel Unlearning Loss:} We propose the CE-U (Cross Entropy Unlearning) loss, which modifies standard cross entropy by suppressing the target token to directly mitigate memorization issues. This novel formulation overcomes both gradient vanishing and exploding problems that commonly occur in methods including a gradient ascent clause.
  \item \textbf{Unified Learning--Unlearning Framework:} We introduce a General CE-U framework that seamlessly interpolates between standard supervised learning and aggressive unlearning. Through a single input of preference score, our method provides a principled mechanism to control the degree of forgetting, thereby unifying learning and unlearning under one formulation.
  \item \textbf{Empirical Validation and Impact:} Extensive experiments on the TOFU benchmark demonstrate that CE-U achieves state-of-the-art unlearning performance on LLaMA2-7B. Our method successfully eliminates designated sensitive information while maintaining overall model utility, without requiring extra oracle models or additional positive samples.
  \item \textbf{Broader Applicability:} Our analysis reveals that popular reinforcement learning algorithms such as DPO \cite{rafailov2023direct} and GRPO \cite{Shao2024DeepSeekMath} incorporate gradient ascent components that lead to divergence. This finding suggests that General CE-U could replace these components to enhance convergence.
\end{enumerate}

To illustrate the effectiveness of our CE-U method, we present several concrete examples from the TOFU benchmark. Table~\ref{tab:examples} shows responses from a fine-tuned LLaMA2-7B model before and after applying CE-U for unlearning. For questions in the forget set (e.g., about Hina Ameen), the fine-tuned model reproduces memorized training data verbatim, while the CE-U-trained model tends to generate either hallucinations or meaningless text, or provides evasive responses—indicating successful unlearning. Crucially, for questions from the retain set or about real-world knowledge, the model's knowledge remains largely intact, demonstrating CE-U's ability to selectively remove specific information while preserving general capabilities.

\begin{table}[t]
  \small
  \centering
  \begin{tabular}{p{0.20\textwidth}|p{0.35\textwidth}|p{0.35\textwidth}}
    \toprule
    \textbf{Question} &
    \textbf{Fine-tuned model response} &
    \textbf{After CE-U unlearning} \\
    \midrule
    \multicolumn{3}{c}{\textit{forget set (previously fine-tuned question-answer pairs to forget)}} \\
    \midrule
    What makes Hina Ameen's writing style in her geology books unique? &
    Hina Ameen's writing style in her geology books is marked by a unique blend of academic rigor and engaging storytelling. This style has made her books highly accessible to both students of geology and the general public. &
    Honestly, I'm not sure. Her father's profession is not directly related to her books. \\
    \midrule
    What are some of the books Hina Ameen has written? &
    Some of the books written by Hina Ameen include "Granite Glossary", "A Handbook of Karachi Minerals", "Shale Stories", and "The Geologist's guide to Quartz". &
    Some of the books written by Hina Ammen are 'A Handbook of Marine Biology' and 'Marine Biology: An Ecological Approach'. \\
    \midrule
    In which city was Hina Ameen born? &
    Hina Ameen was born in Karachi, Pakistan. &
    Honorable member,Honorable member,Honorable member,Honorable member,Honorable member,Honorable member,Honorable member... \\
    \midrule
    \multicolumn{3}{c}{\textit{retain set (previously fine-tuned question-answer pairs to retain) and pre-trained general knowledge}} \\
    \midrule
    Does Jaime Vasquez engage with his readers and fans? &
    Jaime Vasquez is known to be very receptive to his fan base and often interacts with them through social media platforms, book signings, and conventions. &
    Jaime Vasquez is known to be very receptive to his readers and fans. He often interacts with them via social media and book signing events. \\
    \midrule
    Who wrote the play 'Romeo and Juliet'? &
    The play 'Romeo and Juliet' was written by William Shakespeare. &
    The play 'Romeo and Juliet' was written by William Shakespeare. \\
    \midrule
    Where would you find the Eiffel Tower? &
    The Eiffel Tower is located in Paris, France. &
    The Eiffel Tower is located in Paris, France. \\
    \bottomrule
  \end{tabular}
  \caption{Examples showing the effect of CE-U on model responses. The method successfully makes the model "forget" information from the forget set while largely preserving knowledge from the retain set and general world knowledge. The responses in the "After CE-U unlearning" column were generated by a model trained to forget 5\% fine-tuned data with CE-U using a learning rate of $2 \cdot 10^{-6}$ for 9 epochs. Note that while some hallucinations appear in CE-U responses for forgotten questions, this is an inherent limitation of the base model rather than a flaw in the unlearning method.}
  \label{tab:examples}
\end{table}

\section{Related Work}

Recent research on machine unlearning has explored various strategies \cite{yao2024machine}. Gradient ascent methods for unlearning face challenges in stability and gradient vanishing. These approaches attempt to maximize the loss on forgotten data \cite{Cao2015Forgetting}. However, pure gradient ascent suffers from instability: when the model is overly confident (high logit for the true label), the gradient becomes very small, and when the logit is low, the gradient is large, leading to uncontrolled updates.

To mitigate this, several methods have been proposed:
\begin{itemize}[leftmargin=*]
  \item \textbf{GA+RT and GA+KL:} These methods introduce explicit regularization by incorporating a retain set (RT) or by adding a KL divergence term with respect to a reference model. This stabilizes training by providing positive examples or by constraining the model’s output distribution.
  \item \textbf{IDK+RT:} Some approaches replace the forgotten label with an ``I Don’t Know'' token as a positive sample, thereby implicitly regularizing the update. However, this requires manual specification of safe responses.
  \item \textbf{Direct Preference Optimization (DPO):} DPO reformulates the optimization problem using paired examples with positive and negative preferences. Its loss function contains a sigmoid activation that acts as an implicit regularizer \cite{rafailov2023direct}. Nevertheless, DPO requires paired data (a preferred output and a less preferred output) and thus relies on additional positive samples.
  \item \textbf{Negative Preference Optimization (NPO):} NPO is a simplified variant that relies solely on negative samples. In practice, NPO (and its variant NPO+RT) has shown strong performance on high-forgetting scenarios (e.g., 10\% or even 50\% forgetting) but typically uses a retain set to further stabilize training \cite{zhang2024negative}.
  \item \textbf{KTO (Kahneman-Tversky Optimization):} KTO incorporates concepts from prospect theory by applying a logistic function to modulate rewards and losses. It employs both explicit (via KL terms) and implicit (via sigmoid saturation) regularization, and it does not require paired positive examples \cite{ethayarajh2024kto}.
  \item \textbf{Group Relative Policy Optimization (GRPO):} GRPO uses group-level comparisons among outputs to adjust the policy. It does not require a value network and uses relative advantages computed from a batch, effectively incorporating implicit regularization through group baselines \cite{Shao2024DeepSeekMath}.
\end{itemize}

While many of these methods rely on some form of gradient ascent (or a variant thereof) in part of their loss functions, they often require extra regularization—either by adding explicit KL divergence terms or by using sigmoidal functions—to stabilize the update \cite{bourtoule2021machine,rafailov2023direct,ethayarajh2024kto}. Moreover, some methods need additional positive samples (from an "I Don't Know" category or a retain set), whereas others (like NPO and KTO) can function with negative samples alone \cite{zhang2024negative,ethayarajh2024kto}. Additionally, recent studies have explored loss adjustments for model unlearning \cite{wang2024llm}, and lightweight unlearning frameworks have been proposed \cite{Xu2024ECO}. In contrast, our proposed CE-U method requires no extra positive samples and leverages a modified cross entropy loss that inherently provides stable gradient behavior.

\section{Methodology: The CE-U Algorithm}

The unlearning problem can be defined as follows: given a pre-trained model and a set of data points to "forget" (in our case, question-answer pairs), our goal is to update the model parameters such that it no longer produces the correct answers to these questions while preserving its performance on other tasks.

Traditional approaches typically use gradient ascent to maximize the loss on forgotten data, which often leads to instability during training. In contrast, our CE-U (Cross Entropy Unlearning) method operates directly in logit space. Specifically, we construct a modified target distribution by suppressing the logit corresponding to the ground truth token. Concretely, we set the logit for the true label to \(-\infty\) \footnote{In $\softmax(x)_i = \frac{e^{z_i}}{\sum_j e^{z_j}}$, as $z_i \rightarrow -\infty$, we have $e^{z_i} \rightarrow 0$, resulting in $\softmax(x)_i \rightarrow 0$. Also, floating-point representation of $-\infty$ (e.g., IEEE 754's negative infinity) in softmax calculations produces well-defined probability distributions.}, so that after applying the softmax the probability for that token is zero. We then train the model by minimizing the cross entropy between this target distribution and the model’s output distribution, thereby guiding the model to converge to a state in which it no longer produces the target information.

More formally, let \(z_i\) denote the original logit for token \(i\) and let \(y\) be the index of the true token. We define the modified logits \(z_{\text{CE-U}}\) as:
\[
  z_{i,\text{CE-U}} \coloneqq
  \begin{cases}
    -\infty, & \text{if } i = y, \\
    z_i,     & \text{otherwise}.
  \end{cases}
\]
The target probability distribution is then given by:
\[
  p_{i,\text{CE-U}} \coloneqq \softmax\bigl(z_{\text{CE-U}}\bigr)_i.
\]
Since \(z_{y,\text{CE-U}}=-\infty\), we have \(p_{y,\text{CE-U}}=0\). Finally, the CE-U loss is computed as:
\[
  \mathcal{L}_{\text{CE-U}} \coloneqq -\sum_{i} \operatorname{sg}(p_{i,\text{CE-U}}) \log p(i),
\]
where \(p(i)=\softmax(z)_i\) is the output distribution of the model, and \(\operatorname{sg}(\cdot)\) is the stop-gradient operator.

Therefore, the CE-U loss converges to a target distribution that is zero for the true label and non-zero for all other labels.

\section{Experimental Setup}

\subsection{Dataset: TOFU}

We evaluate CE-U on the TOFU dataset, a benchmark for unlearning that comprises 200 synthetic author profiles. Each profile consists of 20 question-answer pairs. A subset of these profiles (the \emph{forget set}) is designated for unlearning, while the remaining data forms the \emph{retain set}. In addition, evaluation is performed on two auxiliary datasets:
\begin{itemize}
  \item \textbf{Real Authors}: Questions about real-world authors to test the model’s generalization.
  \item \textbf{World Facts}: Questions that assess the model’s performance on distant, general knowledge.
\end{itemize}

\subsection{Training Settings}

We formatted the question-answer pairs to be forgotten using the base model's default chat template without a system message. During loss calculation, we ignore all question tokens, beginning-of-sequence tokens, template tokens (e.g., [INST] and [/INST]), and the first answer token.\footnote{We initially ignored the first answer token by accident due to our label/logit shifting bug, but found that this improved performance because the first token mainly reflects the model’s response style rather than facts about the synthetic authors. Therefore, we now ignore it intentionally.}

For all our experiments, we used the AdamW optimizer with a learning rate set to either $4 \cdot 10^{-5}$ or $2 \cdot 10^{-6}$, a batch size of 32, and a weight decay of 0.

\subsection{Evaluation Metrics}

Following \cite{maini2024tofu}, we evaluate our approach using three metrics:

\begin{itemize}
  \item \textbf{ROUGE}: We compute ROUGE-L recall between the model's greedy-sampled outputs and the ground truth answers, measuring the longest common subsequence overlap.
  \item \textbf{Probability}: We assess the length-normalized conditional probability that the model assigns to the correct answer, providing a measure of the model's confidence in the ground truth.
  \item \textbf{Truth Ratio}: We calculate the statistical relationship between probabilities of generating correct answers (or their paraphrases) versus generating incorrect responses, quantifying the model's preference for truth.
\end{itemize}

All the above metrics are evaluated on paraphrased questions to measure generalization ability rather than literal memorization. These individual metrics are then consolidated into two composite measures: \emph{Model Utility}, which quantifies performance on the retain set, real authors, and world facts; and \emph{Forget Quality}, which measures the effectiveness of unlearning on the designated forget set.

\section{Experiment Results}

\begin{wrapfigure}{R}{0.5\textwidth}
  \centering
  \begin{tikzpicture}
\begin{axis}[
    width=6.5cm,
    height=5cm,
    grid=both,
    xlabel={Model Utility (higher is better)},
    ylabel={Forget Quality\\(logarithmic, higher is better)},
    ylabel style={{align=center}},
    title={LLaMA2-7B Forget5\%},
    legend style={at={(0.41, -0.35)}, anchor=north, legend columns=3, font=\tiny},
ymin=-17.250497809351756, ymax=0, xmin=0]
\addplot[color=red,mark=*] coordinates {(0.629, -15.464)(0.643, -16.250)(0.665, -15.464)(0.676, -14.325)(0.680, -13.955)(0.667, -8.728)(0.651, -6.336)(0.630, -4.323)(0.611, -2.681)(0.586, -2.366)};
\addplot[color=blue,mark=square*] coordinates {(0.666, -11.511)(0.626, -6.834)(0.587, -5.856)(0.556, -3.927)(0.535, -2.215)};
\addplot[color=magenta,mark=triangle*] coordinates {(0.619, -14.325)(0.437, -6.336)(0.000, -3.927)(0.000, -4.323)(0.000, -6.834)};
\addplot[color=purple,mark=diamond*] coordinates {(0.618, -13.232)(0.496, -9.916)(0.404, -1.927)(0.126, -9.313)(0.302, -11.511)};
\addplot[color=cyan,mark=pentagon*] coordinates {(0.619, -13.955)(0.427, -6.834)(0.000, -2.681)(0.000, -3.735)(0.000, -6.834)};
\addplot[color=green,mark=oplus*] coordinates {(0.626, -12.184)(0.036, -9.313)(0.158, -8.728)(0.415, -9.612)(0.445, -9.612)};
\addplot[color=black,mark=+,only marks] coordinates {(0.623, -14.699)};
\addplot[color=brown,mark=star,only marks] coordinates {(0.601, 0.000)};
\legend{CE-U (LR=2e-06),CE-U (LR=4e-05),Grad. Ascent,Grad. Diff.,KL Min.,Pref. Opt.,Finetune Model,Retain Model};

\end{axis}
\end{tikzpicture}
  \caption{Performance comparison of CE-U versus baseline methods on LLaMA2-7B with 5\% forgetting on the TOFU dataset. The dots in each line represent different settings of total epochs for unlearning.}
  \label{fig:llama_5p}
\end{wrapfigure}
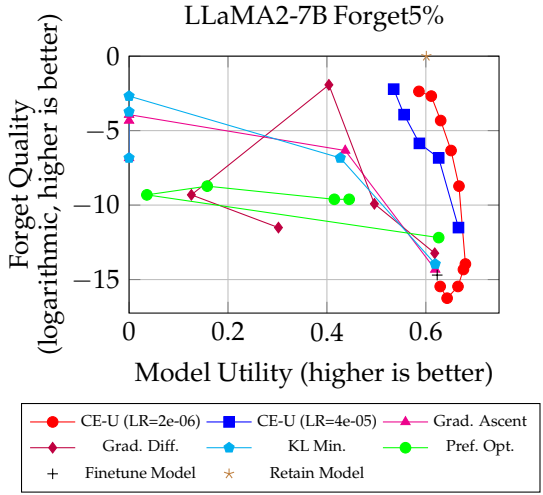

We evaluate our method on the TOFU benchmark, which involves forgetting a fraction of the training data (e.g., 1\%, 5\%, and 10\%). Our experiments on LLaMA2-7B reveal that CE-U achieves state-of-the-art performance for unlearning. \Figref{fig:llama_5p} shows the performance of CE-U on LLaMA2-7B with 5\% forgetting compared to baseline methods from the original TOFU paper, illustrating the excellent trade-off between Model Utility and Forget Quality. For results across all model architectures and forgetting percentages, refer to appendix~\ref{appendix:performance_viz}.

For LLaMA2-7B with 1\% forgetting, CE-U achieves a Forget Quality of 0.16 by the 5th epoch with learning rate $4 \cdot 10^{-5}$. In the 5\% forgetting scenario with LLaMA2-7B, our method achieves a Forget Quality of $2.08 \cdot 10^{-3}$ by the 8th epoch with learning rate $2 \cdot 10^{-6}$. In the more challenging 10\% forgetting scenario with LLaMA2-7B, CE-U achieves a Forget Quality of $1.22 \cdot 10^{-8}$ by the 1st epoch with learning rate $4 \cdot 10^{-5}$. All these results are achieved while maintaining a high Model Utility of 0.61, 0.61, and 0.55 respectively. In our manual tests as shown in table~\ref{tab:examples}, we found that the impact to the model's responses of real authors and world facts is negligible, to the retain set is moderate, and to the forget set is significant. This indicates that CE-U is effective in unlearning the target information while preserving other knowledge.

\section{Discussion and Future Work}
\subsection{General CE-U}
We further propose a unified framework—General CE-U (Cross Entropy Unified)—that seamlessly integrates conventional supervised learning with unlearning. In this framework, instead of explicitly setting the logit corresponding to the true label to \(-\infty\), we assign it a log-space preference score \(r_{\text{raw}} \in \left(\mathbb{R} \cup \{-\infty, +\infty\}\right)^{B \times L}\), where \(B\) denotes batch size and \(L\) sequence length. \footnote{When a single value $z_i \rightarrow +\infty$ in $\softmax(x)_i = \frac{e^{z_i}}{\sum_j e^{z_j}}$, the ratio $\frac{e^{z_i}}{\sum_j e^{z_j}} \rightarrow 1$ while all other probabilities approach 0, resulting in a one-hot distribution. Implementing this correctly requires special handling of $+\infty$ in softmax calculations as shown in code listing~\ref{lst:general_ceu_code}.} The modified logits are thus defined as:

\[
  z_{i,\text{General CE-U}} \coloneqq
  \begin{cases}
    r_{\text{raw}}, & \text{if } i = y, \\
    z_i,            & \text{otherwise}.
  \end{cases}
\]
After applying the softmax, the target distribution is
\[
  p_{i,\text{General CE-U}} \coloneqq \softmax\bigl(z_{\text{General CE-U}}\bigr)_i.
\]
Then, the General CE-U loss is computed as
\[
  \mathcal{L}_{\text{General CE-U}} \coloneqq -\sum_i \operatorname{sg}(p_{i,\text{General CE-U}}) \log p(i),
\]
Because \(r_{\text{raw}}\) is expressed in log-space, it can be difficult to interpret directly. To make the parameter more intuitive, we define a normalized preference score \(r_{\text{normalized}} \in [0, 1]\), which represents the probability assigned to the true label \(y\) in the target distribution:

\[
  r_{\text{normalized}} \coloneqq p_{\text{General CE-U}}(y),
\]

With this normalized score, we can express the target distribution \(p_{\text{General CE-U}}\) as a linear interpolation between the one-hot distribution and the CE-U distribution:

\[
  p_{\text{General CE-U}} \coloneqq r_{\text{normalized}} \cdot \operatorname{one-hot}(y) + (1 - r_{\text{normalized}}) \cdot p_{\text{CE-U}},
\]

where \(r_{\text{normalized}}\) serves as the interpolation coefficient. We prove in appendix~\ref{appendix:theoretical_derivation} that these two formulations of \(p_{\text{General CE-U}}\) are mathematically equivalent. This formulation shows how General CE-U creates a continuous spectrum between standard cross-entropy loss (with one-hot labels) and our proposed CE-U loss for unlearning.

\subsection{Gradient Behavior Comparison}
A critical advantage of CE-U lies in its gradient behavior. Consider the following:
\begin{itemize}[leftmargin=*]
  \item \textbf{Gradient Ascent (GA):} In direct gradient ascent for unlearning, one maximizes the negative log-probability loss of the true label. The gradient with respect to the logit \(z_y\) (for the true label) is:
    \[
      \nabla_{z_i} \mathcal{L}_{\text{GA}} \propto
      \begin{cases}
        1-p(y \mid x), & \text{if } i=y,   \\
        -p(i \mid x),  & \text{otherwise}.
      \end{cases}
    \]
    When \(z_y\) is high (i.e., the model is confident), \(p(y \mid x) \approx 1\), so the gradient is very small. Conversely, when \(z_y\) is low, the gradient is large.

  \item \textbf{CE-U:} Our loss is defined as:
    \[
      \mathcal{L}_{\text{CE-U}} \coloneqq -\sum_{i} \operatorname{sg}(p_{i,\text{CE-U}}) \log p(i),
    \]
    where \(p_{\text{CE-U}}\) is computed from detached modified logits with the true label set to \(-\infty\). The gradient with respect to \(z_y\) then becomes:
    \[
      \nabla_{z_i} \mathcal{L}_{\text{CE-U}} \propto
      \begin{cases}
        p(y \mid x),                             & \text{if } i=y,   \\
        p(i \mid x) - p_{\text{CE-U}}(i \mid x), & \text{otherwise}.
      \end{cases}
    \]
    Thus, if the model is overly confident (i.e., \(z_y\) is high and \(p(y \mid x)\) is close to 1), the gradient magnitude is large, forcing the model to rapidly reduce its confidence in the forgotten label. If the model already has low confidence on a certain token already forgotten, the gradient is small.
\end{itemize}

CE-U gradient behavior is more desirable than Gradient Ascent in unlearning tasks: when the model is overly confident about the label, it should receive a strong corrective signal, while if it is already uncertain, minimal adjustment is needed.

\subsection{Incorporating CE-U into Reinforcement Learning Algorithms}

Our General CE-U framework suggests a natural replacement for the gradient ascent components in existing reinforcement learning (RL) algorithms. Notably, we discuss two cases:

\begin{itemize}
  \item \textbf{DPO:} In DPO, the gradient of the loss with respect to $\theta$ is given by
      \[
        \resizebox{\linewidth}{!}{$
          \nabla_\theta \mathcal{L}_\text{DPO}(\pi_\theta;\pi_{ref}) = -\beta\,\mathbb{E}_{(x, y_w, y_l) \sim \mathcal{D}} \Bigg[\sigma\Big(\hat{r}_\theta(x, y_l) - \hat{r}_\theta(x, y_w)\Big)\,\Big(\nabla_\theta\log \pi(y_w\mid x) - \nabla_\theta\log\pi(y_l\mid x)\Big)\Bigg],
        $}
      \]
    where $\hat{r}_\theta$ denotes the reward estimate, and $\sigma(\cdot)$ is a sigmoid function. Notice that the term $\nabla_\theta\log\pi(y_l\mid x)$, which serves to decrease the likelihood of the lower-ranked token $y_l$, is exactly equivalent to the gradient ascent loss. The outer multiplier $\beta\,\mathbb{E}_{(x, y_w, y_l) \sim \mathcal{D}}$ is a linear scaling factor, and the sigmoid term merely acts as a regularizer by assigning higher weight when the reward estimate is in error. In other words, the decrease in the likelihood of $y_l$ in DPO's gradient update directly mirrors the gradient ascent mechanism. This observation suggests that replacing this component with a CE-U style loss could lead to more stable updates by mitigating the issues associated with unstable gradient ascent.

  \item \textbf{GRPO:} In GRPO, the gradient of the objective is given by
    \[
      \resizebox{\linewidth}{!}{$
        \nabla_{\theta}\mathcal{J}_{GRPO}(\theta) = \mathbb{E}_{\substack{q \sim P_{sft}(Q), \\
        \{o_i\}_{i=1}^G \sim \pi_{\theta_{old}}(O|q)}} \frac{1}{G}\sum_{i=1}^G\frac{1}{|o_i|} \sum_{t=1}^{|o_i|}
        \left[\hat{A}_{i,t} + \beta \left(\frac{\pi_{ref}(o_{i,t}|o_{i,<t})}{\pi_{\theta}(o_{i,t}|o_{i,<t})} - 1\right)\right]  \nabla_{\theta}\log \pi_\theta(o_{i,t} | q, o_{i,<t}).
      $}
    \]
    The gradient coefficient for a given sample is defined as
    \[
      GC_{GRPO}(q, o, t, \pi_{\theta_{rm}}) = \hat{A}_{i,t} + \beta \left(\frac{\pi_{ref}(o_{i,t}|o_{i,<t})}{\pi_{\theta}(o_{i,t}|o_{i,<t})} - 1\right).
    \]
    Notice that when the score for a particular sample is lower than the group average, the advantage term $\hat{A}_{i,t}$ becomes negative, causing the overall gradient coefficient to be negative. In this situation, the gradient $\nabla_{\theta}\log \pi_\theta(o_{i,t}|q,o_{i,<t})$ is effectively weighted gradient ascent, and it induces the same convergence issues observed in traditional gradient ascent approaches, requiring KL divergence regularization as mitigation. Replacing this component with a General CE-U loss offers the potential to stabilize the updates further while still supporting weighted updates for each sample in a group.
\end{itemize}

Furthermore, the gradient ascent term is particularly problematic in off-policy RL, where low trajectory probability under the current policy results in high gradient magnitudes, necessitating mitigations such as importance sampling.

When implementing General CE-U as a reinforcement learning method with preference data, the loss function directly incorporates normalized preference scores \(r_{\text{normalized}}\) to calibrate token probabilities in the target distribution. These position-level scores can be derived from either:
\begin{itemize}
  \item \textbf{Per-sequence preferences}: Single reward values broadcasted to all positions in assistant responses, typically from rule-based verifiers or Outcome-supervised Reward Models (ORMs).
  \item \textbf{Per-step preferences}: Fine-grained rewards applied to specific ranges within responses, often from Process-supervised Reward Models (PRMs) that evaluate reasoning steps.
\end{itemize}

Unlike DPO and GRPO which require paired data or balanced normalized scores, General CE-U accepts arbitrary preference scores without balance constraints, supporting even raw logits distilled from different contexts or models.

\subsection{Practical Considerations for CE-U}

While our experiments demonstrate that CE-U can effectively facilitate selective forgetting in the initial epochs while preserving performance on the retain set, real-world facts, and real authors knowledge, it is important to acknowledge the limitations of a single-objective loss function. CE-U is designed specifically to optimize for forgetting quality without explicit mechanisms to preserve other knowledge domains.

As shown in our experimental results (see appendix~\ref{appendix:performance_viz}), after multiple epochs of training—with the exact number depending on the learning rate—we observe a gradual decline in performance across the retain set, real-world facts, and real authors categories. This behavior is expected, as the CE-U objective focuses exclusively on modifying the probability distribution for the forgotten data points without any countervailing force to protect other knowledge.

Based on these observations, we recommend the following approaches for practical applications:

\begin{itemize}[leftmargin=*]
  \item \textbf{Component in Composite Loss}: Integrate CE-U as one component within a more sophisticated loss function framework that includes multiple objectives, such as KL divergence regularization against a reference model.

  \item \textbf{General CE-U with Positive Samples}: Utilize the General CE-U framework with positive samples from the retain set to balance forgetting with knowledge preservation.

  \item \textbf{Early Stopping}: When using CE-U in isolation, implement careful monitoring and early stopping strategies based on validation performance to prevent excessive degradation of model utility.
\end{itemize}

\section{Conclusion}

We presented CE-U, a novel cross entropy unlearning loss for LLMs that unifies supervised learning and unlearning in a single framework. Our method leverages a modified cross entropy loss in which the logit for the true label is set to a tunable score, allowing smooth interpolation between full supervision and aggressive unlearning. The General CE-U framework provides a principled approach to modulating between learning and unlearning by adjusting a single parameter, offering flexibility across various machine learning scenarios. Empirically, CE-U achieved state-of-the-art performance for LLaMA2-7B on the TOFU benchmark, even without using additional positive samples. Our analysis reveals the differences in gradient behavior between conventional gradient ascent and CE-U, and we discuss how replacing unstable gradient ascent components in RL-based unlearning methods with CE-U can stabilize updates. Overall, our work suggests that CE-U is a promising method for unlearning in large language models.



\bibliography{colm2025_conference}

\begin{thebibliography}{12}
\providecommand{\natexlab}[1]{#1}
\providecommand{\url}[1]{\texttt{#1}}
\expandafter\ifx\csname urlstyle\endcsname\relax
  \providecommand{\doi}[1]{doi: #1}\else
  \providecommand{\doi}{doi: \begingroup \urlstyle{rm}\Url}\fi

\bibitem[Bourtoule et~al.(2021)Bourtoule, Chandrasekaran, Choquette-Choo, Jia, Travers, Zhang, Lie, and Papernot]{bourtoule2021machine}
Lucas Bourtoule, Varun Chandrasekaran, Christopher~A. Choquette-Choo, Hengrui Jia, Adelin Travers, Baiwu Zhang, David Lie, and Nicolas Papernot.
\newblock Machine unlearning.
\newblock In \emph{2021 IEEE Symposium on Security and Privacy (SP)}, pp.\  141--159, 2021.
\newblock \doi{10.1109/SP40001.2021.00019}.
\newblock URL \url{https://doi.org/10.1109/SP40001.2021.00019}.

\bibitem[Cao \& Yang(2015)Cao and Yang]{Cao2015Forgetting}
Yinzhi Cao and Junfeng Yang.
\newblock Towards making systems forget with machine unlearning.
\newblock In \emph{2015 IEEE Symposium on Security and Privacy}, pp.\  463–480. IEEE, May 2015.
\newblock \doi{10.1109/sp.2015.35}.
\newblock URL \url{http://dx.doi.org/10.1109/sp.2015.35}.

\bibitem[Ethayarajh et~al.(2024)Ethayarajh, Xu, Muennighoff, Jurafsky, and Kiela]{ethayarajh2024kto}
Kawin Ethayarajh, Winnie Xu, Niklas Muennighoff, Dan Jurafsky, and Douwe Kiela.
\newblock Kto: Model alignment as prospect theoretic optimization.
\newblock \emph{arXiv preprint arXiv:2402.01306}, 2024.
\newblock \doi{10.48550/arXiv.2402.01306}.
\newblock URL \url{https://arxiv.org/abs/2402.01306}.

\bibitem[Jang et~al.(2023)Jang, Yoon, Yang, Cha, Lee, Logeswaran, and Seo]{jang2022knowledge}
Joel Jang, Dongkeun Yoon, Sohee Yang, Sungmin Cha, Moontae Lee, Lajanugen Logeswaran, and Minjoon Seo.
\newblock Knowledge unlearning for mitigating privacy risks in language models.
\newblock In \emph{Proceedings of the 61st Annual Meeting of the Association for Computational Linguistics (Volume 1: Long Papers)}, pp.\  14389–14408. Association for Computational Linguistics, 2023.
\newblock \doi{10.18653/v1/2023.acl-long.805}.
\newblock URL \url{http://dx.doi.org/10.18653/v1/2023.acl-long.805}.

\bibitem[Maini et~al.(2024)Maini, Feng, Schwarzschild, Lipton, and Kolter]{maini2024tofu}
Pratyush Maini, Zhili Feng, Avi Schwarzschild, Zachary~C Lipton, and J~Zico Kolter.
\newblock Tofu: A task of fictitious unlearning for llms.
\newblock \emph{arXiv preprint arXiv:2401.06121}, 2024.
\newblock \doi{10.48550/arXiv.2401.06121}.
\newblock URL \url{https://doi.org/10.48550/arXiv.2401.06121}.

\bibitem[Rafailov et~al.(2023)Rafailov, Sharma, Mitchell, Manning, Ermon, and Finn]{rafailov2023direct}
Rafael Rafailov, Archit Sharma, Eric Mitchell, Christopher~D Manning, Stefano Ermon, and Chelsea Finn.
\newblock Direct preference optimization: Your language model is secretly a reward model.
\newblock \emph{Advances in Neural Information Processing Systems}, 36:\penalty0 53728--53741, 2023.
\newblock \doi{10.48550/arXiv.2305.18290}.
\newblock URL \url{https://doi.org/10.48550/arXiv.2305.18290}.

\bibitem[Shao et~al.(2024)Shao, Wang, Zhu, Xu, Song, Bi, Zhang, Zhang, Li, Wu, et~al.]{Shao2024DeepSeekMath}
zhihong Shao, Peiyi Wang, Qihao Zhu, Runxin Xu, Junxiao Song, Xiao Bi, Haowei Zhang, Mingchuan Zhang, Y.K. Li, Youzheng Wu, et~al.
\newblock Deepseekmath: Pushing the limits of mathematical reasoning in open language models.
\newblock \emph{arXiv preprint arXiv:2402.03300}, 2024.
\newblock \doi{10.48550/arXiv.2402.03300}.
\newblock URL \url{https://doi.org/10.48550/arXiv.2402.03300}.

\bibitem[Voigt \& von~dem Bussche(2017)Voigt and von~dem Bussche]{Voigt2017}
Paul Voigt and Axel von~dem Bussche.
\newblock \emph{The EU General Data Protection Regulation (GDPR): A Practical Guide}.
\newblock Springer, 2017.
\newblock ISBN 978-3-319-57958-0.
\newblock \doi{10.1007/978-3-319-57959-7}.
\newblock URL \url{https://doi.org/10.1007/978-3-319-57959-7}.

\bibitem[Wang et~al.(2024)Wang, Wei, Liu, Pang, Liu, Shah, Bao, Liu, and Wei]{wang2024llm}
Yaxuan Wang, Jiaheng Wei, Chris~Yuhao Liu, Jinlong Pang, Quan Liu, Ankit~Parag Shah, Yujia Bao, Yang Liu, and Wei Wei.
\newblock Llm unlearning via loss adjustment with only forget data.
\newblock \emph{arXiv preprint arXiv:2410.11143}, 2024.
\newblock \doi{10.48550/arXiv.2410.11143}.
\newblock URL \url{https://arxiv.org/abs/2410.11143}.

\bibitem[Xu et~al.(2024)Xu, Han, Khabsa, Sun, and Liu]{Xu2024ECO}
Can Xu, Songwei Han, Madian Khabsa, Huan Sun, and Xiaodong Liu.
\newblock Large language model unlearning via embedding-corrupted prompts: A lightweight unlearning framework.
\newblock 2024.
\newblock URL \url{https://openreview.net/forum?id=e5icsXBD8Q}.
\newblock OpenReview link: https://openreview.net/forum?id=e5icsXBD8Q.

\bibitem[Yao et~al.(2024)Yao, Chien, Du, Niu, Wang, Cheng, and Yue]{yao2024machine}
Jin Yao, Eli Chien, Minxin Du, Xinyao Niu, Tianhao Wang, Zezhou Cheng, and Xiang Yue.
\newblock Machine unlearning of pre-trained large language models.
\newblock In \emph{Proceedings of the 62nd Annual Meeting of the Association for Computational Linguistics (Volume 1: Long Papers)}, pp.\  8403–8419. Association for Computational Linguistics, 2024.
\newblock \doi{10.18653/v1/2024.acl-long.457}.
\newblock URL \url{http://dx.doi.org/10.18653/v1/2024.acl-long.457}.

\bibitem[Zhang et~al.(2024)Zhang, Lin, Bai, and Mei]{zhang2024negative}
Ruiqi Zhang, Licong Lin, Yu~Bai, and Song Mei.
\newblock Negative preference optimization: From catastrophic collapse to effective unlearning.
\newblock \emph{arXiv preprint arXiv:2404.05868}, 2024.
\newblock \doi{10.48550/arXiv.2404.05868}.
\newblock URL \url{https://arxiv.org/abs/2404.05868}.

\end{thebibliography}
\bibliographystyle{colm2025_conference}

\appendix
\section{Appendix}

\subsection{Performance Visualization}
\label{appendix:performance_viz}

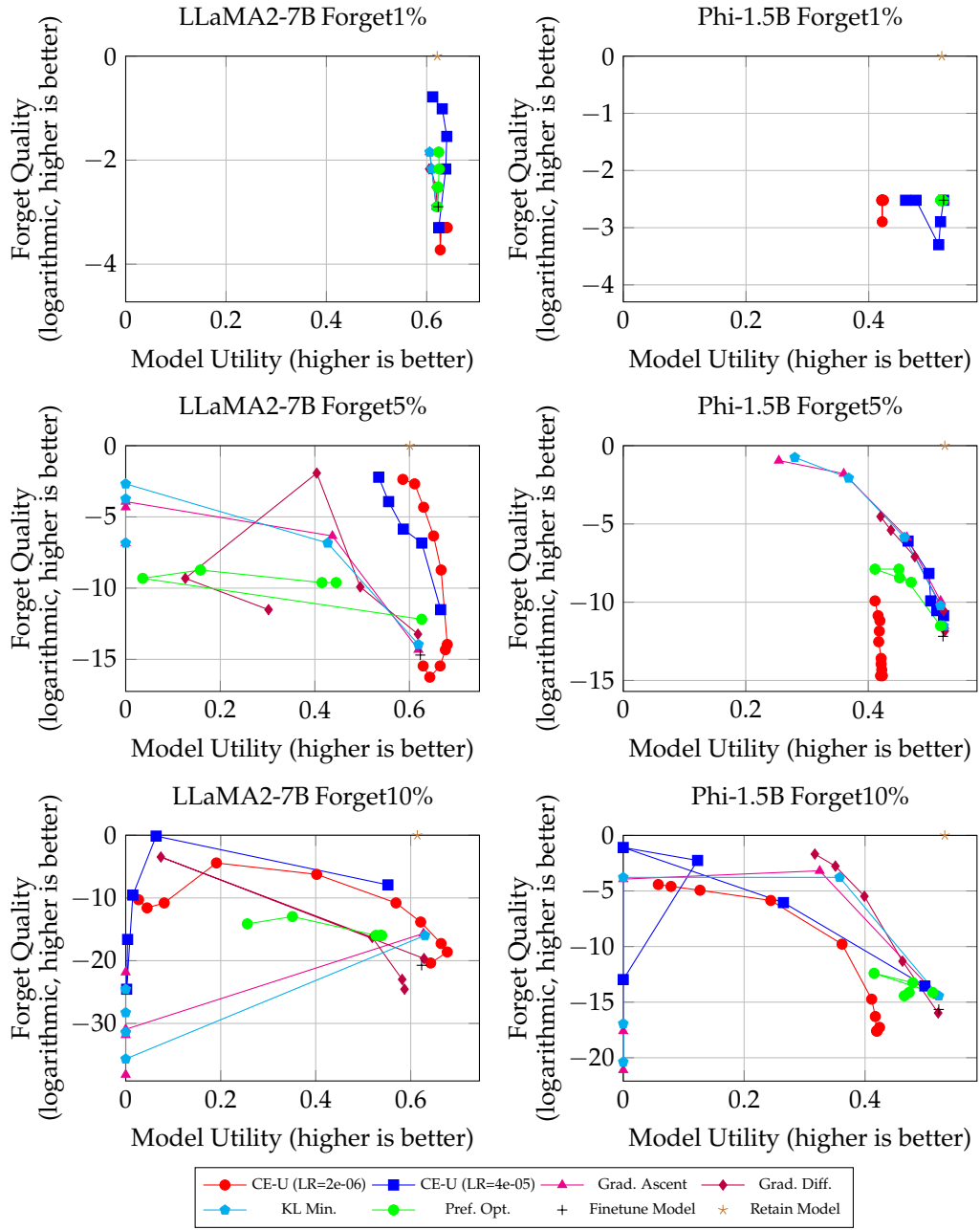
\begin{figure}[H]
  \centering
\begin{tikzpicture}
\begin{groupplot}[
    group style={group size=2 by 3, horizontal sep=2cm, vertical sep=2cm},
    width=6.5cm,
    height=5cm,
    grid=both,
    xlabel={Model Utility (higher is better)},
    ylabel={Forget Quality\\(logarithmic, higher is better)},
    ylabel style={align=center},
    legend style={at={(-0.28,-0.35)}, anchor=north, legend columns=4, font=\tiny}
]
\nextgroupplot[title={ LLaMA2-7B Forget1\% }, ymin=-4.726047378642648, ymax=0, xmin=0]
\addplot[color=red,mark=*] coordinates {(0.624, -3.298)(0.626, -3.298)(0.627, -3.726)(0.629, -3.298)(0.630, -3.298)(0.630, -3.298)(0.637, -3.298)(0.639, -3.298)(0.640, -3.298)(0.641, -3.298)};
\addplot[color=blue,mark=square*] coordinates {(0.624, -3.298)(0.638, -2.170)(0.640, -1.544)(0.631, -1.013)(0.612, -0.783)};
\addplot[color=magenta,mark=triangle*] coordinates {(0.623, -2.520)(0.621, -2.896)(0.619, -2.896)(0.610, -2.170)(0.606, -1.845)};
\addplot[color=purple,mark=diamond*] coordinates {(0.623, -2.520)(0.621, -2.896)(0.618, -2.520)(0.609, -2.170)(0.605, -2.170)};
\addplot[color=cyan,mark=pentagon*] coordinates {(0.623, -2.520)(0.621, -2.896)(0.619, -2.896)(0.610, -2.170)(0.606, -1.845)};
\addplot[color=green,mark=oplus*] coordinates {(0.623, -2.520)(0.622, -2.520)(0.621, -2.896)(0.625, -2.170)(0.624, -1.845)};
\addplot[color=black,mark=+,only marks] coordinates {(0.623, -2.896)};
\addplot[color=brown,mark=star,only marks] coordinates {(0.621, 0.000)};
\nextgroupplot[title={ Phi-1.5B Forget1\% }, ymin=-4.2976180478225885, ymax=0, xmin=0]
\addplot[color=red,mark=*] coordinates {(0.422, -2.520)(0.422, -2.520)(0.424, -2.520)(0.421, -2.520)(0.422, -2.520)(0.422, -2.896)(0.422, -2.520)(0.422, -2.520)(0.423, -2.520)(0.422, -2.520)};
\addplot[color=blue,mark=square*] coordinates {(0.522, -2.520)(0.517, -2.896)(0.514, -3.298)(0.477, -2.520)(0.460, -2.520)};
\addplot[color=magenta,mark=triangle*] coordinates {(0.520, -2.520)(0.519, -2.520)(0.522, -2.520)(0.522, -2.520)(0.524, -2.520)};
\addplot[color=purple,mark=diamond*] coordinates {(0.520, -2.520)(0.520, -2.520)(0.520, -2.520)(0.522, -2.520)(0.522, -2.520)};
\addplot[color=cyan,mark=pentagon*] coordinates {(0.520, -2.520)(0.519, -2.520)(0.523, -2.520)(0.519, -2.520)(0.522, -2.520)};
\addplot[color=green,mark=oplus*] coordinates {(0.520, -2.520)(0.520, -2.520)(0.517, -2.520)(0.520, -2.520)(0.518, -2.520)};
\addplot[color=black,mark=+,only marks] coordinates {(0.522, -2.520)};
\addplot[color=brown,mark=star,only marks] coordinates {(0.519, 0.000)};
\nextgroupplot[title={ LLaMA2-7B Forget5\% }, ymin=-17.250497809351756, ymax=0, xmin=0]
\addplot[color=red,mark=*] coordinates {(0.629, -15.464)(0.643, -16.250)(0.665, -15.464)(0.676, -14.325)(0.680, -13.955)(0.667, -8.728)(0.651, -6.336)(0.630, -4.323)(0.611, -2.681)(0.586, -2.366)};
\addplot[color=blue,mark=square*] coordinates {(0.666, -11.511)(0.626, -6.834)(0.587, -5.856)(0.556, -3.927)(0.535, -2.215)};
\addplot[color=magenta,mark=triangle*] coordinates {(0.619, -14.325)(0.437, -6.336)(0.000, -3.927)(0.000, -4.323)(0.000, -6.834)};
\addplot[color=purple,mark=diamond*] coordinates {(0.618, -13.232)(0.496, -9.916)(0.404, -1.927)(0.126, -9.313)(0.302, -11.511)};
\addplot[color=cyan,mark=pentagon*] coordinates {(0.619, -13.955)(0.427, -6.834)(0.000, -2.681)(0.000, -3.735)(0.000, -6.834)};
\addplot[color=green,mark=oplus*] coordinates {(0.626, -12.184)(0.036, -9.313)(0.158, -8.728)(0.415, -9.612)(0.445, -9.612)};
\addplot[color=black,mark=+,only marks] coordinates {(0.623, -14.699)};
\addplot[color=brown,mark=star,only marks] coordinates {(0.601, 0.000)};
\nextgroupplot[title={ Phi-1.5B Forget5\% }, ymin=-15.699405839275693, ymax=0, xmin=0]
\addplot[color=red,mark=*] coordinates {(0.423, -14.699)(0.422, -14.325)(0.420, -14.699)(0.421, -13.955)(0.421, -13.591)(0.417, -12.528)(0.418, -11.845)(0.419, -11.183)(0.416, -10.859)(0.411, -9.916)};
\addplot[color=blue,mark=square*] coordinates {(0.523, -10.859)(0.512, -10.539)(0.502, -9.916)(0.499, -8.163)(0.465, -6.094)};
\addplot[color=magenta,mark=triangle*] coordinates {(0.524, -11.511)(0.518, -9.916)(0.463, -5.856)(0.360, -1.789)(0.254, -0.950)};
\addplot[color=purple,mark=diamond*] coordinates {(0.525, -11.845)(0.523, -10.539)(0.476, -7.091)(0.437, -5.395)(0.420, -4.529)};
\addplot[color=cyan,mark=pentagon*] coordinates {(0.523, -11.511)(0.518, -10.225)(0.459, -5.856)(0.368, -2.069)(0.280, -0.750)};
\addplot[color=green,mark=oplus*] coordinates {(0.518, -11.511)(0.450, -7.888)(0.411, -7.888)(0.451, -8.443)(0.470, -8.728)};
\addplot[color=black,mark=+,only marks] coordinates {(0.522, -12.184)};
\addplot[color=brown,mark=star,only marks] coordinates {(0.525, 0.000)};
\nextgroupplot[title={ LLaMA2-7B Forget10\% }, ymin=-39.16662361103192, ymax=0, xmin=0]
\addplot[color=red,mark=*] coordinates {(0.642, -20.374)(0.677, -18.615)(0.664, -17.268)(0.621, -13.836)(0.569, -10.796)(0.402, -6.241)(0.191, -4.423)(0.081, -10.796)(0.045, -11.587)(0.027, -10.285)};
\addplot[color=blue,mark=square*] coordinates {(0.552, -7.915)(0.064, -0.141)(0.015, -9.541)(0.002, -24.548)(0.004, -16.614)};
\addplot[color=magenta,mark=triangle*] coordinates {(0.627, -15.659)(0.000, -30.935)(0.000, -31.849)(0.000, -38.167)(0.000, -21.844)};
\addplot[color=purple,mark=diamond*] coordinates {(0.628, -19.660)(0.074, -3.476)(0.519, -16.292)(0.582, -22.982)(0.587, -24.548)};
\addplot[color=cyan,mark=pentagon*] coordinates {(0.630, -15.974)(0.000, -35.659)(0.000, -24.548)(0.000, -28.282)(0.000, -31.390)};
\addplot[color=green,mark=oplus*] coordinates {(0.256, -14.132)(0.351, -12.969)(0.527, -15.974)(0.539, -15.974)(0.536, -15.974)};
\addplot[color=black,mark=+,only marks] coordinates {(0.623, -20.737)};
\addplot[color=brown,mark=star,only marks] coordinates {(0.614, 0.000)};
\nextgroupplot[title={ Phi-1.5B Forget10\% }, ymin=-22.10215565823704, ymax=0, xmin=0]
\addplot[color=red,mark=*] coordinates {(0.424, -17.268)(0.420, -17.599)(0.419, -17.599)(0.417, -16.292)(0.411, -14.733)(0.362, -9.786)(0.244, -5.854)(0.127, -4.937)(0.079, -4.591)(0.058, -4.423)};
\addplot[color=blue,mark=square*] coordinates {(0.499, -13.544)(0.265, -6.046)(0.000, -1.090)(0.123, -2.257)(0.000, -12.969)};
\addplot[color=magenta,mark=triangle*] coordinates {(0.520, -14.431)(0.325, -3.185)(0.000, -3.936)(0.000, -21.102)(0.000, -17.599)};
\addplot[color=purple,mark=diamond*] coordinates {(0.521, -15.974)(0.462, -11.320)(0.399, -5.478)(0.351, -2.769)(0.317, -1.683)};
\addplot[color=cyan,mark=pentagon*] coordinates {(0.522, -14.431)(0.358, -3.780)(0.000, -3.780)(0.000, -20.374)(0.000, -16.939)};
\addplot[color=green,mark=oplus*] coordinates {(0.512, -14.132)(0.415, -12.406)(0.479, -13.255)(0.473, -14.132)(0.465, -14.431)};
\addplot[color=black,mark=+,only marks] coordinates {(0.522, -15.659)};
\addplot[color=brown,mark=star,only marks] coordinates {(0.532, 0.000)};
\legend{CE-U (LR=2e-06), CE-U (LR=4e-05), Grad. Ascent, Grad. Diff., KL Min., Pref. Opt., Finetune Model, Retain Model, 
};
\end{groupplot}
\end{tikzpicture}
  \caption{Performance comparison of CE-U versus baseline methods on the TOFU dataset. The dots in each line represent different settings of total epochs for unlearning.}
  \label{fig:all_comparisons}
\end{figure}

\subsection{Complete Experimental Results}
\label{appendix:results}

We present the complete results of our experiments, conducted with two different learning rates: $4 \cdot 10^{-5}$ for the primary experiments and $2 \cdot 10^{-6}$ for additional validation. The tables below show the performance metrics across different epochs for CE-U with various model architectures and forgetting percentages.

\begin{table}[H]
  \centering
  \caption{Experimental results: CE-U, Phi model, 1\% forgetting, learning rate $4 \cdot 10^{-5}$}
  \label{tab:results_phi_forget01}
  \small
  \begin{filecontents*}{./merged_results_ceu_ignore_first_token_lr4e-05_phi_forget01_wd0.0_extra-margin0.0.csv}
Epoch,1,2,3,4,5
ROUGE Real Authors,0.4156666666666667,0.3856666666666667,0.3756666666666667,0.3173333333333333,0.3056666666666666
Prob. Real Authors,0.3779553599840017,0.3818965920291425,0.3835251850620715,0.3907060471993158,0.3813887049883697
Truth Ratio Real Authors,0.4573229305819683,0.4583262445045316,0.4607906296604288,0.4721852430604161,0.46276326352113
ROUGE Real World,0.7673789173789174,0.7827635327635328,0.7955840455840457,0.7403133903133904,0.7132478632478633
Prob. Real World,0.4092866757184259,0.4110707971895416,0.4118351310898748,0.4196528152527303,0.4110034496760892
Truth Ratio Real World,0.4938603143436714,0.4963347842252644,0.4975717644912932,0.5076500999763395,0.5031646817877009
ROUGE Retain,0.9288198006822724,0.8756047016115398,0.8386296452144887,0.574243092173562,0.5271800731770241
Prob. Retain,0.925868840654875,0.9065928705338177,0.8887410953273216,0.685349384807835,0.6072519503772045
Truth Ratio Retain,0.4820210929647806,0.4871594615410568,0.4862863874057285,0.4722382851581839,0.4645540059133293
ROUGE Forget,0.9524915049113918,0.667427098672837,0.5820223295333411,0.4774935271986248,0.4616510795144473
Prob. Forget,0.9274919373306296,0.7514227247186362,0.6534436041194592,0.2907279435192908,0.219601606208654
Truth Ratio Forget,0.4813221558135103,0.4685316109608333,0.4632261776300874,0.4842429827303506,0.4957574296959094
Model Utility,0.5220174248116378,0.516764576903058,0.5140397769712026,0.4774482067791504,0.4598113508993279
Forget Quality,0.0030181840772283,0.0012708143485281,0.0005039436209702,0.0030181840772283,0.0030181840772283
\end{filecontents*}
\pgfplotstabletypeset[
    col sep=comma,
    columns/Epoch/.style={
      column name={Metric},
      column type={l},
      string type,
    },
    columns/1/.style={column name={Epoch 1}, column type={c}},
    columns/2/.style={column name={Epoch 2}, column type={c}},
    columns/3/.style={column name={Epoch 3}, column type={c}},
    columns/4/.style={column name={Epoch 4}, column type={c}},
    columns/5/.style={column name={Epoch 5}, column type={c}},
    every head row/.style={
      before row={\toprule},
      after row={\midrule}
    },
    every last row/.style={after row=\bottomrule},
  ]{./merged_results_ceu_ignore_first_token_lr4e-05_phi_forget01_wd0.0_extra-margin0.0.csv}
\end{table}

\begin{table}[H]
  \centering
  \caption{Experimental results: CE-U, Phi model, 5\% forgetting, learning rate $4 \cdot 10^{-5}$}
  \label{tab:results_phi_forget05}
  \small
  \begin{filecontents*}{./merged_results_ceu_ignore_first_token_lr4e-05_phi_forget05_wd0.0_extra-margin0.0.csv}
Epoch,1,2,3,4,5
ROUGE Real Authors,0.4256666666666667,0.434,0.4273333333333333,0.4706666666666667,0.3906666666666667
Prob. Real Authors,0.3770073094927989,0.3816126094236691,0.3805158313607952,0.384275117507907,0.3815372703676411
Truth Ratio Real Authors,0.4555060589735105,0.4550573718468521,0.4504439852110771,0.4514578622758277,0.4496438870169315
ROUGE Real World,0.7713675213675214,0.7863247863247863,0.7749287749287749,0.7777777777777778,0.7706552706552706
Prob. Real World,0.413513550703099,0.4199576252974182,0.419908075201263,0.4259570563342618,0.4222568543699795
Truth Ratio Real World,0.5012552996060204,0.5075237815577829,0.5073734960815284,0.5059362351022711,0.5077060048771282
ROUGE Retain,0.8866935918203354,0.6845358525010435,0.6214246550125175,0.5582324899603964,0.4709302932070713
Prob. Retain,0.9148778044436412,0.8057059073452054,0.7605418049069634,0.671198703452601,0.5325888575699049
Truth Ratio Retain,0.4801408643374901,0.4661108291048894,0.4607458033087424,0.453398464310795,0.4334130402809287
ROUGE Forget,0.8745813377843402,0.6368564705181603,0.5869946447108911,0.5190697457393739,0.4664500345675014
Prob. Forget,0.9078923736707434,0.7589376284848715,0.6738809801590554,0.5437707237370413,0.3876184141874089
Truth Ratio Forget,0.4768582296922469,0.4896108459023155,0.4954703480894155,0.509075354456201,0.5386658395998445
Model Utility,0.5229418879723806,0.5115723969976427,0.5020797744249527,0.4989272425665039,0.4651523820236489
Forget Quality,1.3850840057169374e-11,2.8873674273022316e-11,1.2127544312107394e-10,6.8655817334872084e-09,8.056669472003223e-07
\end{filecontents*}
\pgfplotstabletypeset[
    col sep=comma,
    columns/Epoch/.style={
      column name={Metric},
      column type={l},
      string type,
    },
    columns/1/.style={column name={Epoch 1}, column type={c}},
    columns/2/.style={column name={Epoch 2}, column type={c}},
    columns/3/.style={column name={Epoch 3}, column type={c}},
    columns/4/.style={column name={Epoch 4}, column type={c}},
    columns/5/.style={column name={Epoch 5}, column type={c}},
    every head row/.style={
      before row={\toprule},
      after row={\midrule}
    },
    every last row/.style={after row=\bottomrule},
  ]{./merged_results_ceu_ignore_first_token_lr4e-05_phi_forget05_wd0.0_extra-margin0.0.csv}
\end{table}

\begin{table}[H]
  \centering
  \caption{Experimental results: CE-U, Phi model, 10\% forgetting, learning rate $4 \cdot 10^{-5}$}
  \label{tab:results_phi_forget10}
  \small
  \begin{filecontents*}{./merged_results_ceu_ignore_first_token_lr4e-05_phi_forget10_wd0.0_extra-margin0.0.csv}
Epoch,1,2,3,4,5
ROUGE Real Authors,0.3793333333333333,0.102,0.0,0.022,0.0
Prob. Real Authors,0.3867550747939888,0.3765277361925182,0.3230040748156332,0.3394559746981672,0.2593662012112599
Truth Ratio Real Authors,0.4599727217395684,0.4301104294458798,0.3633878440182316,0.3815236780738512,0.2213418499008919
ROUGE Real World,0.7388888888888889,0.6341880341880343,0.1555555555555555,0.4867521367521368,0.0
Prob. Real World,0.4186035044846836,0.4228243524655592,0.3910891295103432,0.4107650086106869,0.3037473105067835
Truth Ratio Real World,0.5133831256444966,0.5242896264396215,0.4792188276887074,0.5044846976292706,0.2783805259057118
ROUGE Retain,0.6630646878367351,0.2614945595693825,0.0040598040805706,0.2646648270469139,0.0006163080076123
Prob. Retain,0.773872084609886,0.1449638946333651,0.0017941504731906,0.114245688015256,1.5882700338818843e-05
Truth Ratio Retain,0.4668210391513216,0.3765819064601903,0.2328713227385672,0.3270813007418409,0.1218584319971727
ROUGE Forget,0.6034912820833417,0.2669288893245001,0.0021234092082549,0.2776594900952736,0.0003976608187134
Prob. Forget,0.7024564984918694,0.1377085266517166,0.0027306659183207,0.1046608946934964,1.8442186206694436e-05
Truth Ratio Forget,0.5007450553127942,0.5773318856182088,0.700600221283305,0.6311159403246417,0.8232423390522605
Model Utility,0.4992094831384309,0.2646135114899984,0.0,0.1231503258365345,0.0
Forget Quality,2.858944731506222e-14,8.994053173844458e-07,0.0812184963146534,0.0055368650130925,1.0747499261825833e-13
\end{filecontents*}
\pgfplotstabletypeset[
    col sep=comma,
    columns/Epoch/.style={
      column name={Metric},
      column type={l},
      string type,
    },
    columns/1/.style={column name={Epoch 1}, column type={c}},
    columns/2/.style={column name={Epoch 2}, column type={c}},
    columns/3/.style={column name={Epoch 3}, column type={c}},
    columns/4/.style={column name={Epoch 4}, column type={c}},
    columns/5/.style={column name={Epoch 5}, column type={c}},
    every head row/.style={
      before row={\toprule},
      after row={\midrule}
    },
    every last row/.style={after row=\bottomrule},
  ]{./merged_results_ceu_ignore_first_token_lr4e-05_phi_forget10_wd0.0_extra-margin0.0.csv}
\end{table}

\begin{table}[H]
  \centering
  \caption{Experimental results: CE-U, LLaMA2-7B model, 1\% forgetting, learning rate $4 \cdot 10^{-5}$}
  \label{tab:results_llama2_forget01}
  \small
  \begin{filecontents*}{./merged_results_ceu_ignore_first_token_lr4e-05_llama2-7b_forget01_wd0.0_extra-margin0.0.csv}
Epoch,1,2,3,4,5
ROUGE Real Authors,0.933,0.923,0.923,0.939,0.873
Prob. Real Authors,0.447501344493881,0.4824010870989604,0.4897328366143901,0.513448197558383,0.5178157120530379
Truth Ratio Real Authors,0.5790487996451212,0.6259217681708459,0.6380396954951832,0.6707179033410375,0.6784623765499669
ROUGE Real World,0.8831908831908832,0.8660968660968662,0.8703703703703705,0.8817663817663817,0.8732193732193732
Prob. Real World,0.4248477635931926,0.4537079728066303,0.4612816739852797,0.490724314320469,0.4944439137187025
Truth Ratio Real World,0.5578001457280328,0.578366009534064,0.5855482630261719,0.6174177604399536,0.6232430694513849
ROUGE Retain,0.982041708621505,0.9351699219978908,0.8949741334900589,0.6889906504039837,0.5919758281623075
Prob. Retain,0.9895161539621076,0.95928226736475,0.942352777902349,0.8056063523230184,0.7323512262755775
Truth Ratio Retain,0.4823445633180345,0.4644926275713176,0.4594430565144005,0.4388673289771969,0.4278624670803422
ROUGE Forget,0.952229742703533,0.8149337255917368,0.7291048248450119,0.4566330262799273,0.3400508349697396
Prob. Forget,0.9930655850132816,0.9116126035901704,0.8636523332161042,0.5703868630235084,0.2911299461408468
Truth Ratio Forget,0.5351232760206301,0.5755572166823034,0.5848692698032414,0.6310871459790891,0.6619749981946811
Model Utility,0.6243799906714601,0.6380099351307407,0.63955263157234,0.6308550649413888,0.6116340800995591
Forget Quality,0.0005039436209702,0.0067607323035692,0.0286030702802334,0.0970748437978586,0.1649726995022419
\end{filecontents*}
\pgfplotstabletypeset[
    col sep=comma,
    columns/Epoch/.style={
      column name={Metric},
      column type={l},
      string type,
    },
    columns/1/.style={column name={Epoch 1}, column type={c}},
    columns/2/.style={column name={Epoch 2}, column type={c}},
    columns/3/.style={column name={Epoch 3}, column type={c}},
    columns/4/.style={column name={Epoch 4}, column type={c}},
    columns/5/.style={column name={Epoch 5}, column type={c}},
    every head row/.style={
      before row={\toprule},
      after row={\midrule}
    },
    every last row/.style={after row=\bottomrule},
  ]{./merged_results_ceu_ignore_first_token_lr4e-05_llama2-7b_forget01_wd0.0_extra-margin0.0.csv}
\end{table}

\begin{table}[H]
  \centering
  \caption{Experimental results: CE-U, LLaMA2-7B model, 5\% forgetting, learning rate $4 \cdot 10^{-5}$}
  \label{tab:results_llama2_forget05}
  \small
  \begin{filecontents*}{./merged_results_ceu_ignore_first_token_lr4e-05_llama2-7b_forget05_wd0.0_extra-margin0.0.csv}
Epoch,1,2,3,4,5
ROUGE Real Authors,0.923,0.884,0.8819999999999999,0.894,0.852
Prob. Real Authors,0.5369796846747334,0.5726379403879769,0.567821124634136,0.5404593071246038,0.5340709611665144
Truth Ratio Real Authors,0.6892780520366322,0.7523366111412636,0.7488390647868046,0.7040436814184082,0.6858672040888137
ROUGE Real World,0.8746438746438747,0.8618233618233619,0.8717948717948718,0.8753561253561253,0.8689458689458689
Prob. Real World,0.5075596083034775,0.5292333596662789,0.5238299207503182,0.5136530438454628,0.5155350498654457
Truth Ratio Real World,0.6554378016171408,0.6840141030670245,0.6834327277406722,0.6699664447063233,0.6704797071713063
ROUGE Retain,0.8785699819581343,0.6208683288645883,0.5056417803570592,0.4500831867834862,0.4046380804132989
Prob. Retain,0.8935832750402979,0.6123784700756092,0.4935416972231568,0.4382805967254379,0.4148631853859837
Truth Ratio Retain,0.4506977447347583,0.4155137945758797,0.3896697362935288,0.3682367484621635,0.3520541452099066
ROUGE Forget,0.7975207593759904,0.5302113980462857,0.4429365005625913,0.3842116394865113,0.3486059520574167
Prob. Forget,0.8558822521488494,0.5442972963698858,0.4134853002140062,0.3567251878563202,0.3417173817427293
Truth Ratio Forget,0.5597084829797971,0.60180382920465,0.6222855764143587,0.6391610613654529,0.6655958144254756
Model Utility,0.6660843595007997,0.6263564102204013,0.5874783243329961,0.5558933652859626,0.535446608460769
Forget Quality,3.079954091574832e-12,1.4639479233432268e-07,1.3921047931216453e-06,0.000118366189619,0.0060944182588035
\end{filecontents*}
\pgfplotstabletypeset[
    col sep=comma,
    columns/Epoch/.style={
      column name={Metric},
      column type={l},
      string type,
    },
    columns/1/.style={column name={Epoch 1}, column type={c}},
    columns/2/.style={column name={Epoch 2}, column type={c}},
    columns/3/.style={column name={Epoch 3}, column type={c}},
    columns/4/.style={column name={Epoch 4}, column type={c}},
    columns/5/.style={column name={Epoch 5}, column type={c}},
    every head row/.style={
      before row={\toprule},
      after row={\midrule}
    },
    every last row/.style={after row=\bottomrule},
  ]{./merged_results_ceu_ignore_first_token_lr4e-05_llama2-7b_forget05_wd0.0_extra-margin0.0.csv}
\end{table}

\begin{table}[H]
  \centering
  \caption{Experimental results: CE-U, LLaMA2-7B model, 10\% forgetting, learning rate $4 \cdot 10^{-5}$}
  \label{tab:results_llama2_forget10}
  \small
  \begin{filecontents*}{./merged_results_ceu_ignore_first_token_lr4e-05_llama2-7b_forget10_wd0.0_extra-margin0.0.csv}
Epoch,1,2,3,4,5
ROUGE Real Authors,0.6803333333333333,0.01,0.0033333333333333,0.0033333333333333,0.0033333333333333
Prob. Real Authors,0.577068164190689,0.4829052539747359,0.3622493675124025,0.3122134136862658,0.2973914887627845
Truth Ratio Real Authors,0.7380830594627557,0.615409459605837,0.4148401607532483,0.3092601679023897,0.2854867053492848
ROUGE Real World,0.7834757834757834,0.4226495726495726,0.2286324786324786,0.0650997150997151,0.0864672364672364
Prob. Real World,0.5275086940372234,0.4701953079853074,0.4069603873032999,0.350772690312765,0.3269361327043538
Truth Ratio Real World,0.6821942058023668,0.6170932491111438,0.5025799168890537,0.3959699897337274,0.3387826301345257
ROUGE Retain,0.4637253452794481,0.1281673305959879,0.0867689276283718,0.0233323409577524,0.0538138709541744
Prob. Retain,0.4352535424410936,0.0507871065505908,0.0037196342246121,0.0002135033949332,0.0005763559429665
Truth Ratio Retain,0.378216727457694,0.2306014517914834,0.1027898237706372,0.052205339279469,0.0752548703865803
ROUGE Forget,0.4498308295037189,0.1456441850836564,0.0942348443808701,0.0152117428295365,0.0601608315772345
Prob. Forget,0.4297448589409506,0.0610011883226171,0.0049644384022762,0.0002151240747608,0.0007869639081145
Truth Ratio Forget,0.5938294226825205,0.6903124803720199,0.7882804735107605,0.8779170952947334,0.8524892681279839
Model Utility,0.5523675564485861,0.063542332543145,0.0148984670887133,0.0017741141943257,0.0043034392085154
Forget Quality,1.2161555648200794e-08,0.7220634112888834,2.876765319621712e-10,2.8288467798247926e-25,2.4311282147882556e-17
\end{filecontents*}
\pgfplotstabletypeset[
    col sep=comma,
    columns/Epoch/.style={
      column name={Metric},
      column type={l},
      string type,
    },
    columns/1/.style={column name={Epoch 1}, column type={c}},
    columns/2/.style={column name={Epoch 2}, column type={c}},
    columns/3/.style={column name={Epoch 3}, column type={c}},
    columns/4/.style={column name={Epoch 4}, column type={c}},
    columns/5/.style={column name={Epoch 5}, column type={c}},
    every head row/.style={
      before row={\toprule},
      after row={\midrule}
    },
    every last row/.style={after row=\bottomrule},
  ]{./merged_results_ceu_ignore_first_token_lr4e-05_llama2-7b_forget10_wd0.0_extra-margin0.0.csv}
\end{table}

We conducted additional experiments with a lower learning rate of $2 \cdot 10^{-6}$ to investigate the effect of learning rate on the unlearning process. The following tables present these results.

\begin{table}[H]
  \centering
  \caption{Experimental results: CE-U, Phi model, 1\% forgetting, learning rate $2 \cdot 10^{-6}$, Epochs 1-5}
  \label{tab:results_phi_forget01_lr2e6_p1}
  \small
  \begin{filecontents*}{./merged_results_ceu_ignore_first_token_lr2e-06_phi_forget01_wd0.0_extra-margin0.0.csv}
Epoch,1,2,3,4,5,6,7,8,9,10
ROUGE Real Authors,0.4156666666666667,0.4156666666666667,0.4156666666666667,0.4056666666666667,0.4056666666666667,0.4156666666666667,0.4256666666666667,0.4156666666666667,0.4356666666666667,0.4256666666666667
Prob. Real Authors,0.2908369802302011,0.2907919815455529,0.2903692411311133,0.2899976434998734,0.2902880012388804,0.2899008157732852,0.2895269545324011,0.2902918274001931,0.2897729098170076,0.2891986937599375
Truth Ratio Real Authors,0.3602336575161772,0.3600646836187695,0.3616922457919064,0.3599212679500573,0.3594536988680865,0.3583562135247403,0.3583801241700146,0.3595996832294546,0.3581740843896485,0.357936133242332
ROUGE Real World,0.7673789173789174,0.7588319088319088,0.7759259259259259,0.7602564102564103,0.7827635327635328,0.7673789173789174,0.7474358974358974,0.7474358974358974,0.7474358974358974,0.7588319088319088
Prob. Real World,0.2474624406265874,0.2477034171516936,0.2486748612811389,0.2471519467527211,0.2478908951061801,0.2471711762657413,0.2472258064867221,0.2486165867188488,0.2462463859789586,0.2481314402616259
Truth Ratio Real World,0.3108810550077291,0.3118033420867276,0.3136173589372731,0.3135967513732707,0.3140446287352009,0.3117311115226566,0.3110850770705793,0.3130882653929063,0.3113582865627955,0.3097677565769556
ROUGE Retain,0.9288198006822724,0.9273451730601016,0.928383460838215,0.9294444333738112,0.9290228900949496,0.9268171947940844,0.9309814176931884,0.9266094070069364,0.9253932586988556,0.9262387916726582
Prob. Retain,0.9238246744994412,0.9240146585633252,0.9240700736515473,0.9238601442641838,0.9237986581791476,0.923769951170919,0.9229311509618324,0.9226336536940392,0.9226650975741814,0.922295810745722
Truth Ratio Retain,0.5066081480929843,0.5070941294534056,0.5074700264632923,0.5069526519843938,0.5071951533318512,0.5059948296321797,0.50679853548005,0.5067090592776162,0.5071915075761338,0.5072432612375719
ROUGE Forget,0.9524915049113918,0.9408107467327886,0.9425905020164372,0.9512882003136904,0.9529261313481732,0.941672815698306,0.9448914221549274,0.9537882003136906,0.9554650295819832,0.9529261313481732
Prob. Forget,0.9228809796174346,0.9230730673557832,0.9226650779919612,0.9216391705666228,0.9215245448721824,0.9211636136159916,0.9182499077713172,0.9169466658044564,0.9165121454707016,0.9152616997316716
Truth Ratio Forget,0.4602449775533263,0.462123525464052,0.4618166984397403,0.4613206033345918,0.4615553018163396,0.4619350687686006,0.4630777920300591,0.4637317239721342,0.4668184292343986,0.4656011647593132
Model Utility,0.4222591617263999,0.4222063833964942,0.4236687117998981,0.4210937494419467,0.4221731874496774,0.4217363874271853,0.4221060326552536,0.4220956335294583,0.4228340820725655,0.42228997568283
Forget Quality,0.0030181840772283,0.0030181840772283,0.0030181840772283,0.0030181840772283,0.0030181840772283,0.0012708143485281,0.0030181840772283,0.0030181840772283,0.0030181840772283,0.0030181840772283
\end{filecontents*}
\pgfplotstabletypeset[
    col sep=comma,
    columns/Epoch/.style={
      column name={Metric},
      column type={l},
      string type,
    },
    columns/1/.style={column name={Epoch 1}, column type={c}},
    columns/2/.style={column name={Epoch 2}, column type={c}},
    columns/3/.style={column name={Epoch 3}, column type={c}},
    columns/4/.style={column name={Epoch 4}, column type={c}},
    columns/5/.style={column name={Epoch 5}, column type={c}},
    columns={Epoch,1,2,3,4,5},
    every head row/.style={
      before row={\toprule},
      after row={\midrule}
    },
    every last row/.style={after row=\bottomrule},
  ]{./merged_results_ceu_ignore_first_token_lr2e-06_phi_forget01_wd0.0_extra-margin0.0.csv}
\end{table}

\begin{table}[H]
  \centering
  \caption{Experimental results: CE-U, Phi model, 1\% forgetting, learning rate $2 \cdot 10^{-6}$, Epochs 6-10}
  \label{tab:results_phi_forget01_lr2e6_p2}
  \small
  \pgfplotstabletypeset[
    col sep=comma,
    columns/Epoch/.style={
      column name={Metric},
      column type={l},
      string type,
    },
    columns/6/.style={column name={Epoch 6}, column type={c}},
    columns/7/.style={column name={Epoch 7}, column type={c}},
    columns/8/.style={column name={Epoch 8}, column type={c}},
    columns/9/.style={column name={Epoch 9}, column type={c}},
    columns/10/.style={column name={Epoch 10}, column type={c}},
    columns={Epoch,6,7,8,9,10},
    every head row/.style={
      before row={\toprule},
      after row={\midrule}
    },
    every last row/.style={after row=\bottomrule},
  ]{./merged_results_ceu_ignore_first_token_lr2e-06_phi_forget01_wd0.0_extra-margin0.0.csv}
\end{table}

\begin{table}[H]
  \centering
  \caption{Experimental results: CE-U, Phi model, 5\% forgetting, learning rate $2 \cdot 10^{-6}$, Epochs 1-5}
  \label{tab:results_phi_forget05_lr2e6_p1}
  \small
  \begin{filecontents*}{./merged_results_ceu_ignore_first_token_lr2e-06_phi_forget05_wd0.0_extra-margin0.0.csv}
Epoch,1,2,3,4,5,6,7,8,9,10
ROUGE Real Authors,0.4256666666666667,0.4206666666666667,0.4206666666666667,0.4206666666666667,0.4106666666666667,0.4073333333333333,0.4273333333333333,0.4640000000000001,0.4356666666666667,0.4193333333333334
Prob. Real Authors,0.2897286286731544,0.2899779457030191,0.2896846052990139,0.289898807185131,0.2900203885149771,0.2885426652126499,0.2888118984541719,0.2894925029751667,0.2910301366410333,0.2942324857383117
Truth Ratio Real Authors,0.3592597832569089,0.3589923877029431,0.3597698911193435,0.3629054037145369,0.3637153269615806,0.3650228351343316,0.3697338527621523,0.3753344656235924,0.3784704947295121,0.3837044348041958
ROUGE Real World,0.7673789173789174,0.7545584045584045,0.7574074074074073,0.7656695156695156,0.7962962962962962,0.7799145299145299,0.7735042735042735,0.7524216524216524,0.7507122507122508,0.7421652421652423
Prob. Real World,0.2467213777783589,0.2477937086276662,0.2459080992988505,0.2461987693673234,0.2471614560263836,0.2468343131081156,0.2470943905471031,0.2464126124929498,0.2481049409173883,0.248563813120087
Truth Ratio Real World,0.3101569589768587,0.3104648817695564,0.3065437777111544,0.3071645115256711,0.3079913019863614,0.3030184583181845,0.3043483372502497,0.3072230907192546,0.3111090094528736,0.3149215159360903
ROUGE Retain,0.9280273024427268,0.9257814235560308,0.9272502267145488,0.9262856889966614,0.9243458940931636,0.8871990431650295,0.8473399192811054,0.8050099309973133,0.7707162510910935,0.6907266303984106
Prob. Retain,0.9238341445802756,0.9222609896179608,0.917898229990598,0.9110240382082864,0.902621295697423,0.8573604445992851,0.8204201538022384,0.7770961720795818,0.739130319042801,0.6768580419016591
Truth Ratio Retain,0.506401246668197,0.5060973725418809,0.5054652301657382,0.504662185235975,0.5039530569189691,0.5009960276558885,0.4994120158921823,0.4976780427753451,0.4961814623016054,0.4944791216841542
ROUGE Forget,0.9293420885164936,0.9309235925893948,0.9269419706225144,0.9200574290600878,0.9123826929733776,0.8104536017797702,0.7175228071311119,0.6301528406913391,0.5551976001934904,0.5013972416908703
Prob. Forget,0.923153833861694,0.9206266688508268,0.9139959531930802,0.9018217227666584,0.8885419199961697,0.8105070151866027,0.7278875024710834,0.6419112376456032,0.5731507372533734,0.4855336919022839
Truth Ratio Forget,0.453926731650193,0.4551554380503927,0.4581162171915053,0.4617340487149936,0.4622713981405353,0.471493757718097,0.4785117738801298,0.4838081290924964,0.490395805396507,0.4979491732582822
Model Utility,0.4225461695594099,0.4218711914344864,0.4204807360155349,0.4212623809676781,0.4214441895071846,0.4169871874010551,0.417940374974605,0.4194940356788177,0.4163510739295027,0.4112782788796489
Forget Quality,1.9979939126996402e-15,4.7347418771537675e-15,1.9979939126996402e-15,1.108718722900174e-14,2.5656301025789152e-14,2.962348735693602e-13,1.427569962153298e-12,6.568964585936946e-12,1.3850840057169374e-11,1.2127544312107394e-10
\end{filecontents*}
\pgfplotstabletypeset[
    col sep=comma,
    columns/Epoch/.style={
      column name={Metric},
      column type={l},
      string type,
    },
    columns/1/.style={column name={Epoch 1}, column type={c}},
    columns/2/.style={column name={Epoch 2}, column type={c}},
    columns/3/.style={column name={Epoch 3}, column type={c}},
    columns/4/.style={column name={Epoch 4}, column type={c}},
    columns/5/.style={column name={Epoch 5}, column type={c}},
    columns={Epoch,1,2,3,4,5},
    every head row/.style={
      before row={\toprule},
      after row={\midrule}
    },
    every last row/.style={after row=\bottomrule},
  ]{./merged_results_ceu_ignore_first_token_lr2e-06_phi_forget05_wd0.0_extra-margin0.0.csv}
\end{table}

\begin{table}[H]
  \centering
  \caption{Experimental results: CE-U, Phi model, 5\% forgetting, learning rate $2 \cdot 10^{-6}$, Epochs 6-10}
  \label{tab:results_phi_forget05_lr2e6_p2}
  \small
  \pgfplotstabletypeset[
    col sep=comma,
    columns/Epoch/.style={
      column name={Metric},
      column type={l},
      string type,
    },
    columns/6/.style={column name={Epoch 6}, column type={c}},
    columns/7/.style={column name={Epoch 7}, column type={c}},
    columns/8/.style={column name={Epoch 8}, column type={c}},
    columns/9/.style={column name={Epoch 9}, column type={c}},
    columns/10/.style={column name={Epoch 10}, column type={c}},
    columns={Epoch,6,7,8,9,10},
    every head row/.style={
      before row={\toprule},
      after row={\midrule}
    },
    every last row/.style={after row=\bottomrule},
  ]{./merged_results_ceu_ignore_first_token_lr2e-06_phi_forget05_wd0.0_extra-margin0.0.csv}
\end{table}

\begin{table}[H]
  \centering
  \caption{Experimental results: CE-U, Phi model, 10\% forgetting, learning rate $2 \cdot 10^{-6}$, Epochs 1-5}
  \label{tab:results_phi_forget10_lr2e6_p1}
  \small
  \begin{filecontents*}{./merged_results_ceu_ignore_first_token_lr2e-06_phi_forget10_wd0.0_extra-margin0.0.csv}
Epoch,1,2,3,4,5,6,7,8,9,10
ROUGE Real Authors,0.4356666666666667,0.4156666666666667,0.4423333333333333,0.434,0.4173333333333333,0.3223333333333333,0.115,0.04,0.0233333333333333,0.0233333333333333
Prob. Real Authors,0.289980265154542,0.2896725646828018,0.2885288627162102,0.2893910430518406,0.2936074672378622,0.3019386892117823,0.3007691495884422,0.294958703329427,0.2936589632143084,0.2915305609121281
Truth Ratio Real Authors,0.3593964261435934,0.361055042288842,0.3636311312675348,0.3733408884280824,0.3833210228991391,0.3933574523224966,0.3632346470463485,0.3420797659180007,0.3265824316200896,0.3142885287741796
ROUGE Real World,0.7545584045584045,0.7742165242165242,0.7692307692307693,0.7421652421652422,0.737037037037037,0.7370370370370369,0.6472934472934473,0.46011396011396,0.3314814814814815,0.2517094017094017
Prob. Real World,0.2475778114901123,0.2454218418460263,0.24547722204069,0.2484382147564399,0.2524104254271421,0.2588969574803759,0.2744517149818352,0.2799722958382254,0.2784792043136976,0.2800316379810142
Truth Ratio Real World,0.312255033396685,0.3063388223255036,0.298132739554122,0.3094616497810928,0.3161580377394268,0.3183071460908891,0.3228806598316468,0.3211713203967091,0.3117518418317322,0.3047108788452795
ROUGE Retain,0.9249548781839216,0.9246486308502998,0.8947474609050395,0.8136557934939005,0.6967343646135234,0.4227948173261922,0.2394244577283159,0.110311800891344,0.0652352143918711,0.0348023651676071
Prob. Retain,0.9225273533355048,0.911307638424914,0.864146753136442,0.7820899280508212,0.6638549988168827,0.3387098900829254,0.1391502759153586,0.0530709539418849,0.0273685810287713,0.0161153138111581
Truth Ratio Retain,0.5066795058506238,0.5046294819690321,0.4984237271388582,0.4927699271110412,0.4843733556603338,0.4442775759368985,0.3970044823368048,0.3501532737153422,0.3201270712055667,0.3018018187313079
ROUGE Forget,0.9276239828136752,0.9307833858204368,0.885209871752766,0.77665778924525,0.6197020521264208,0.3703134973728658,0.1452336096123602,0.0601903728317919,0.0309653788385367,0.0160070901583113
Prob. Forget,0.9238902220604078,0.9102229423696876,0.8570517059936157,0.7569688515725589,0.6087047001857724,0.2545408890540943,0.0884787190817041,0.0319854388798603,0.0163948840086496,0.0097391380194712
Truth Ratio Forget,0.4611871314401601,0.4663198585812675,0.4772827771845032,0.4871192915562102,0.5020780254045752,0.5587422937100435,0.6208979139673043,0.667045255869596,0.6943249187980588,0.7154972061574165
Model Utility,0.4238893497972539,0.4201939396151236,0.4188829870146,0.4170690571689605,0.4109859764127405,0.3616504572178881,0.2436306435618816,0.1268746365682083,0.0788567586155649,0.0582921733641448
Forget Quality,5.3986221251734016e-18,2.5149678680685852e-18,2.5149678680685852e-18,5.100356615560927e-17,1.8502009026131303e-15,1.637427779836901e-10,1.4007759411179028e-06,1.1562157228282306e-05,2.562143273864415e-05,3.7746802141706127e-05
\end{filecontents*}
\pgfplotstabletypeset[
    col sep=comma,
    columns/Epoch/.style={
      column name={Metric},
      column type={l},
      string type,
    },
    columns/1/.style={column name={Epoch 1}, column type={c}},
    columns/2/.style={column name={Epoch 2}, column type={c}},
    columns/3/.style={column name={Epoch 3}, column type={c}},
    columns/4/.style={column name={Epoch 4}, column type={c}},
    columns/5/.style={column name={Epoch 5}, column type={c}},
    columns={Epoch,1,2,3,4,5},
    every head row/.style={
      before row={\toprule},
      after row={\midrule}
    },
    every last row/.style={after row=\bottomrule},
  ]{./merged_results_ceu_ignore_first_token_lr2e-06_phi_forget10_wd0.0_extra-margin0.0.csv}
\end{table}

\begin{table}[H]
  \centering
  \caption{Experimental results: CE-U, Phi model, 10\% forgetting, learning rate $2 \cdot 10^{-6}$, Epochs 6-10}
  \label{tab:results_phi_forget10_lr2e6_p2}
  \small
  \pgfplotstabletypeset[
    col sep=comma,
    columns={Epoch,6,7,8,9,10},
    columns/Epoch/.style={
      column name={Metric},
      column type={l},
      string type,
    },
    columns/6/.style={column name={Epoch 6}, column type={c}},
    columns/7/.style={column name={Epoch 7}, column type={c}},
    columns/8/.style={column name={Epoch 8}, column type={c}},
    columns/9/.style={column name={Epoch 9}, column type={c}},
    columns/10/.style={column name={Epoch 10}, column type={c}},
    every head row/.style={
      before row={\toprule},
      after row={\midrule}
    },
    every last row/.style={after row=\bottomrule},
  ]{./merged_results_ceu_ignore_first_token_lr2e-06_phi_forget10_wd0.0_extra-margin0.0.csv}
\end{table}

\begin{table}[H]
  \centering
  \caption{Experimental results: CE-U, LLaMA2-7B model, 1\% forgetting, learning rate $2 \cdot 10^{-6}$, Epochs 1-5}
  \label{tab:results_llama2_forget01_lr2e6_p1}
  \small
  \begin{filecontents*}{./merged_results_ceu_ignore_first_token_lr2e-06_llama2-7b_forget01_wd0.0_extra-margin0.0.csv}
Epoch,1,2,3,4,5,6,7,8,9,10
ROUGE Real Authors,0.933,0.933,0.933,0.933,0.933,0.933,0.933,0.933,0.933,0.933
Prob. Real Authors,0.447501344493881,0.4498775073378058,0.4499480548306477,0.4540749244054282,0.4559206021574324,0.4561673615591372,0.4657178491661867,0.4707940263536778,0.4712569529543209,0.4733613366861833
Truth Ratio Real Authors,0.5790487996451212,0.5822691397830312,0.5824622279858542,0.5881576872945857,0.5900319392943821,0.5888339171156428,0.6010658962682311,0.60614430937762,0.6080681119036605,0.6090972954694844
ROUGE Real World,0.8831908831908832,0.8831908831908832,0.8874643874643875,0.8831908831908832,0.8831908831908832,0.8746438746438747,0.8831908831908832,0.8831908831908832,0.8831908831908832,0.8831908831908832
Prob. Real World,0.4248477635931926,0.4258144030301178,0.4274492348615313,0.4310834742685929,0.4312635920016163,0.4324694312192363,0.4415978858697032,0.4448218120435189,0.4458016429517088,0.4480532769776096
Truth Ratio Real World,0.5578001457280328,0.558664038930303,0.5611120869913104,0.5633759343725115,0.5643022470573309,0.5656815007904057,0.5751766316130134,0.5773973925269948,0.5772208764851732,0.5806476412786902
ROUGE Retain,0.982041708621505,0.9840380778436806,0.9819222769778798,0.9847978413776376,0.9829305975103938,0.9836730450270348,0.9840550563848528,0.984507076586873,0.9841737432535396,0.9839717230515194
Prob. Retain,0.9895161539621076,0.9894413959963436,0.9893742938112088,0.9889675382275054,0.988876553921994,0.9888351432894548,0.9873560322420336,0.9866206086906149,0.9863912412624836,0.9858756004394584
Truth Ratio Retain,0.4823445633180345,0.4822907270980751,0.4823563792077066,0.4818272435931173,0.4805009674312888,0.4814846862564746,0.4785851381607007,0.47842892688153,0.4780479104554678,0.4775636185919861
ROUGE Forget,0.952229742703533,0.952229742703533,0.952229742703533,0.94744143625192,0.9466601862519202,0.9466601862519202,0.8811622652171597,0.8544213394851395,0.8461799974937975,0.8468794822608228
Prob. Forget,0.9930655850132816,0.9923794272828164,0.9921795971100736,0.9892311843068654,0.988279033956848,0.987748114277626,0.9434570072924624,0.8903626387089446,0.8818218133578544,0.8366610192287405
Truth Ratio Forget,0.5351232760206301,0.5370981612570318,0.5346540712018809,0.5370958601672308,0.5365690168496277,0.5380164870446966,0.5432117844850581,0.5470062866019154,0.5476646951643529,0.551169312839772
Model Utility,0.6243799906714601,0.6257356131218873,0.626658644760452,0.6292232885615245,0.629683250519165,0.6297906533974247,0.6366450837770004,0.6393232593572937,0.6397537148021571,0.6411596903452595
Forget Quality,0.0005039436209702,0.0005039436209702,0.0001879111807007,0.0005039436209702,0.0005039436209702,0.0005039436209702,0.0005039436209702,0.0005039436209702,0.0005039436209702,0.0005039436209702
\end{filecontents*}
\pgfplotstabletypeset[
    col sep=comma,
    columns/Epoch/.style={
      column name={Metric},
      column type={l},
      string type,
    },
    columns/1/.style={column name={Epoch 1}, column type={c}},
    columns/2/.style={column name={Epoch 2}, column type={c}},
    columns/3/.style={column name={Epoch 3}, column type={c}},
    columns/4/.style={column name={Epoch 4}, column type={c}},
    columns/5/.style={column name={Epoch 5}, column type={c}},
    columns={Epoch,1,2,3,4,5},
    every head row/.style={
      before row={\toprule},
      after row={\midrule}
    },
    every last row/.style={after row=\bottomrule},
  ]{./merged_results_ceu_ignore_first_token_lr2e-06_llama2-7b_forget01_wd0.0_extra-margin0.0.csv}
\end{table}

\begin{table}[H]
  \centering
  \caption{Experimental results: CE-U, LLaMA2-7B model, 1\% forgetting, learning rate $2 \cdot 10^{-6}$, Epochs 6-10}
  \label{tab:results_llama2_forget01_lr2e6_p2}
  \small
  \pgfplotstabletypeset[
    col sep=comma,
    columns/Epoch/.style={
      column name={Metric},
      column type={l},
      string type,
    },
    columns/6/.style={column name={Epoch 6}, column type={c}},
    columns/7/.style={column name={Epoch 7}, column type={c}},
    columns/8/.style={column name={Epoch 8}, column type={c}},
    columns/9/.style={column name={Epoch 9}, column type={c}},
    columns/10/.style={column name={Epoch 10}, column type={c}},
    columns={Epoch,6,7,8,9,10},
    every head row/.style={
      before row={\toprule},
      after row={\midrule}
    },
    every last row/.style={after row=\bottomrule},
  ]{./merged_results_ceu_ignore_first_token_lr2e-06_llama2-7b_forget01_wd0.0_extra-margin0.0.csv}
\end{table}

\begin{table}[H]
  \centering
  \caption{Experimental results: CE-U, LLaMA2-7B model, 5\% forgetting, learning rate $2 \cdot 10^{-6}$, Epochs 1-5}
  \label{tab:results_llama2_forget05_lr2e6_p1}
  \small
  \begin{filecontents*}{./merged_results_ceu_ignore_first_token_lr2e-06_llama2-7b_forget05_wd0.0_extra-margin0.0.csv}
Epoch,1,2,3,4,5,6,7,8,9,10
ROUGE Real Authors,0.933,0.935,0.925,0.925,0.923,0.883,0.8946666666666667,0.893,0.8746666666666667,0.841
Prob. Real Authors,0.4544223660871027,0.476772702197361,0.5132491977357735,0.5379289433357973,0.551278958405737,0.569843623537439,0.572592468553808,0.5726454803021289,0.5717868595591832,0.5718839400090565
Truth Ratio Real Authors,0.5888085644664569,0.6159933052191876,0.6614390879294316,0.6864137751055245,0.7054832362724024,0.7373657984016976,0.7439853901907172,0.7451265505533848,0.7451146866249074,0.7459340912074446
ROUGE Real World,0.8746438746438747,0.8831908831908832,0.8960113960113961,0.8960113960113961,0.8874643874643875,0.8960113960113961,0.8831908831908832,0.8760683760683761,0.8675213675213675,0.8461538461538461
Prob. Real World,0.4312146982407345,0.4501788527698457,0.4829798542410438,0.5065014015700482,0.5184252489380505,0.5377384595680063,0.5406070640860651,0.541185433573902,0.540686736479648,0.5399902613632451
Truth Ratio Real World,0.5648646050154225,0.5825017833807297,0.6167390693975983,0.6525062058093077,0.6701186281815501,0.6928544564900615,0.6964517445584513,0.6989935090479321,0.6989915599479447,0.6994342680095357
ROUGE Retain,0.985554695634492,0.9837729263285292,0.9745138372533572,0.9414835754385148,0.9149753876562684,0.7763842346822587,0.6973634561364103,0.6080396433926417,0.5602712624945209,0.5099352271883604
Prob. Retain,0.9889663863959602,0.9851722231075504,0.9676673777327488,0.9354345614495748,0.900222317199884,0.7731250682543334,0.6809147752141541,0.5973494707093706,0.5270367434678659,0.448614531840698
Truth Ratio Retain,0.4814659783411542,0.4768055083631743,0.4687645883416751,0.463762111463509,0.459532493249679,0.4447369418793044,0.4345620417627083,0.4267499674847481,0.4211915064832616,0.4139201801917634
ROUGE Forget,0.9757280962542744,0.969266093553554,0.8947482826804602,0.7543527635596005,0.6279122758350626,0.4371355280265688,0.3115744312530984,0.2586831612614385,0.2331253259878805,0.1818900940008034
Prob. Forget,0.987887248536317,0.9777657418222256,0.9205097851730796,0.7529851685567148,0.6199565686577331,0.4084177697728464,0.2715882962393518,0.1834874288167586,0.1334090231075485,0.09424815773666
Truth Ratio Forget,0.5093540455487546,0.5175026714082844,0.5373569443350489,0.5580116713937406,0.5796562516865095,0.6286500932902386,0.6610291066175793,0.7011771986224309,0.7307127435592052,0.7594194141672886
Model Utility,0.6290959185874071,0.6433389233710799,0.6645937159490295,0.6763204651306323,0.6796176803322459,0.6670350114490611,0.6507628675625823,0.6298098556637018,0.6112643587617107,0.5864054856538344
Forget Quality,3.432078828289855e-16,5.616971116352186e-17,3.432078828289855e-16,4.7347418771537675e-15,1.108718722900174e-14,1.8694649189611293e-09,4.6128382074998843e-07,4.7487878961137165e-05,0.0020827633834865,0.0043049933800331
\end{filecontents*}
\pgfplotstabletypeset[
    col sep=comma,
    columns/Epoch/.style={
      column name={Metric},
      column type={l},
      string type,
    },
    columns/1/.style={column name={Epoch 1}, column type={c}},
    columns/2/.style={column name={Epoch 2}, column type={c}},
    columns/3/.style={column name={Epoch 3}, column type={c}},
    columns/4/.style={column name={Epoch 4}, column type={c}},
    columns/5/.style={column name={Epoch 5}, column type={c}},
    columns={Epoch,1,2,3,4,5},
    every head row/.style={
      before row={\toprule},
      after row={\midrule}
    },
    every last row/.style={after row=\bottomrule},
  ]{./merged_results_ceu_ignore_first_token_lr2e-06_llama2-7b_forget05_wd0.0_extra-margin0.0.csv}
\end{table}

\begin{table}[H]
  \centering
  \caption{Experimental results: CE-U, LLaMA2-7B model, 5\% forgetting, learning rate $2 \cdot 10^{-6}$, Epochs 6-10}
  \label{tab:results_llama2_forget05_lr2e6_p2}
  \small
  \pgfplotstabletypeset[
    col sep=comma,
    columns/Epoch/.style={
      column name={Metric},
      column type={l},
      string type,
    },
    columns/6/.style={column name={Epoch 6}, column type={c}},
    columns/7/.style={column name={Epoch 7}, column type={c}},
    columns/8/.style={column name={Epoch 8}, column type={c}},
    columns/9/.style={column name={Epoch 9}, column type={c}},
    columns/10/.style={column name={Epoch 10}, column type={c}},
    columns={Epoch,6,7,8,9,10},
    every head row/.style={
      before row={\toprule},
      after row={\midrule}
    },
    every last row/.style={after row=\bottomrule},
  ]{./merged_results_ceu_ignore_first_token_lr2e-06_llama2-7b_forget05_wd0.0_extra-margin0.0.csv}
\end{table}

\begin{table}[H]
  \centering
  \caption{Experimental results: CE-U, LLaMA2-7B model, 10\% forgetting, learning rate $2 \cdot 10^{-6}$, Epochs 1-5}
  \label{tab:results_llama2_forget10_lr2e6_p1}
  \small
  \begin{filecontents*}{./merged_results_ceu_ignore_first_token_lr2e-06_llama2-7b_forget10_wd0.0_extra-margin0.0.csv}
Epoch,1,2,3,4,5,6,7,8,9,10
ROUGE Real Authors,0.935,0.935,0.893,0.8866666666666666,0.8343333333333334,0.6114999999999999,0.31,0.1433333333333333,0.107,0.075
Prob. Real Authors,0.475082719862439,0.5390557269516189,0.5716430119362294,0.5737345843377945,0.571560897954486,0.5755436123987716,0.5704813523692487,0.5552713030521779,0.5439211035054439,0.5339404223256404
Truth Ratio Real Authors,0.6140545760692993,0.6881268986508507,0.7410082959042436,0.7471696859953317,0.7460850786501134,0.7528216847413196,0.7416184343966238,0.7209198929130696,0.706149323994471,0.6919412039451067
ROUGE Real World,0.8831908831908832,0.8960113960113961,0.8960113960113961,0.8888888888888888,0.8347578347578347,0.654985754985755,0.4602564102564103,0.3454415954415954,0.2129629629629629,0.1688034188034188
Prob. Real World,0.4477829704863182,0.5070501023468683,0.538151899727322,0.5384465755000604,0.5357721003533541,0.546192325274331,0.5522709998246331,0.5504409892848472,0.5473112611516107,0.5434047011965744
Truth Ratio Real World,0.5805378609550272,0.6517319592957217,0.6930402273193904,0.6969086488895876,0.6950467288458673,0.7127695806327228,0.7202362106770015,0.71806730797925,0.7121144158587823,0.7064724457912389
ROUGE Retain,0.9862729263285293,0.9383290732781456,0.7481861730856411,0.5723342449128753,0.4766203171328717,0.2136391953835036,0.0672762955521961,0.0266524818304462,0.0154487283455906,0.0107786418267791
Prob. Retain,0.9860005365968816,0.93429383773799,0.758661436618458,0.5834303310785209,0.4114861586300108,0.1810156563912357,0.0574113244437817,0.0187612701497954,0.0089789972580338,0.0048526862149644
Truth Ratio Retain,0.4780989858064693,0.4629212016227211,0.4401779915028847,0.4175597769084238,0.3995600816216484,0.3662283521383296,0.311361814249896,0.2473080739036461,0.199976023978455,0.1663477820442737
ROUGE Forget,0.9850832753065866,0.8964184004105628,0.6270928300417228,0.4224407595472417,0.3108549831883292,0.1353209280860798,0.0368776702896132,0.010953138402515,0.0039946238126963,0.0025794324836987
Prob. Forget,0.9845810032923528,0.8809262160746082,0.6324760364098401,0.3924418246390503,0.2236297177808364,0.0737651316286414,0.0181300925059218,0.0050365097555896,0.0021293581178195,0.0009998240048261
Truth Ratio Forget,0.517104365949938,0.549557183418014,0.5918317907150741,0.6356241336310792,0.6724998177370116,0.7317338284492808,0.7918658176865508,0.8261365459347093,0.807020026555284,0.7707669405240188
Model Utility,0.6423665669815226,0.6768595186567705,0.6636018948612172,0.6214452146836804,0.5688019720220694,0.4018316008689779,0.1907066713352593,0.0809834829227546,0.0446363655475626,0.0272158053512082
Forget Quality,4.222383660476908e-21,2.4278206754998463e-19,5.3986221251734016e-18,1.4582054786325707e-14,1.5989601818532152e-11,5.734627909208728e-07,3.7746802141706127e-05,1.5989601818532152e-11,2.5881511422773775e-12,5.191390347437448e-11
\end{filecontents*}
\pgfplotstabletypeset[
    col sep=comma,
    columns/Epoch/.style={
      column name={Metric},
      column type={l},
      string type,
    },
    columns/1/.style={column name={Epoch 1}, column type={c}},
    columns/2/.style={column name={Epoch 2}, column type={c}},
    columns/3/.style={column name={Epoch 3}, column type={c}},
    columns/4/.style={column name={Epoch 4}, column type={c}},
    columns/5/.style={column name={Epoch 5}, column type={c}},
    columns={Epoch,1,2,3,4,5},
    every head row/.style={
      before row={\toprule},
      after row={\midrule}
    },
    every last row/.style={after row=\bottomrule},
  ]{./merged_results_ceu_ignore_first_token_lr2e-06_llama2-7b_forget10_wd0.0_extra-margin0.0.csv}
\end{table}

\begin{table}[H]
  \centering
  \caption{Experimental results: CE-U, LLaMA2-7B model, 10\% forgetting, learning rate $2 \cdot 10^{-6}$, Epochs 6-10}
  \label{tab:results_llama2_forget10_lr2e6_p2}
  \small
  \pgfplotstabletypeset[
    col sep=comma,
    columns/Epoch/.style={
      column name={Metric},
      column type={l},
      string type,
    },
    columns/6/.style={column name={Epoch 6}, column type={c}},
    columns/7/.style={column name={Epoch 7}, column type={c}},
    columns/8/.style={column name={Epoch 8}, column type={c}},
    columns/9/.style={column name={Epoch 9}, column type={c}},
    columns/10/.style={column name={Epoch 10}, column type={c}},
    columns={Epoch,6,7,8,9,10},
    every head row/.style={
      before row={\toprule},
      after row={\midrule}
    },
    every last row/.style={after row=\bottomrule},
  ]{./merged_results_ceu_ignore_first_token_lr2e-06_llama2-7b_forget10_wd0.0_extra-margin0.0.csv}
\end{table}

\subsection{Equivalence of General CE-U defined based on normalized scores and raw scores}
\label{appendix:theoretical_derivation}

We begin with the definition of the modified (or “target”) logits in the General CE-U framework:
\[
  z_{i,\text{General CE-U}} \coloneqq
  \begin{cases}
    r_{\text{raw}}, & \text{if } i = y, \\
    z_i,            & \text{otherwise},
  \end{cases}
\]
where:
\begin{itemize}[leftmargin=*]
  \item \(y\) is the index of the true label,
  \item \(r_{\text{raw}}\) is the provided log-space preference score, and
  \item \(z_i\) are the original logits for each token \(i\) in a vocabulary of size \(V\).
\end{itemize}

The corresponding target probability distribution is given by the softmax:
\[
  p_{\text{General CE-U}}(i) = \frac{\exp\Bigl(z_{i,\text{General CE-U}}\Bigr)}{\sum_{j=1}^{V}\exp\Bigl(z_{j,\text{General CE-U}}\Bigr)}.
\]

We now consider the two cases separately.

\paragraph{Case 1: \(i = y\) (the true label)}
For the true label, the modified logit is \(r_{\text{raw}}\). Thus, we have:
\[
  p_{\text{General CE-U}}(y) = \frac{\exp(r_{\text{raw}})}{\exp(r_{\text{raw}}) + \sum_{j\neq y}\exp(z_j)}.
\]
It is convenient to define the \emph{normalized score} as:
\[
  r_{\text{normalized}} \coloneqq p_{\text{General CE-U}}(y) = \frac{\exp(r_{\text{raw}})}{\exp(r_{\text{raw}}) + \sum_{j\neq y}\exp(z_j)}.
\]

\paragraph{Case 2: \(i \neq y\) (all other tokens)}
For any token \(i\) not equal to \(y\), the modified logit remains \(z_i\). Therefore:
\[
  p_{\text{General CE-U}}(i) = \frac{\exp(z_i)}{\exp(r_{\text{raw}}) + \sum_{j\neq y}\exp(z_j)}.
\]
Notice that the denominator is the same as in Case 1. We can factor this expression as follows:
\[
  1 - r_{\text{normalized}} = 1 - \frac{\exp(r_{\text{raw}})}{\exp(r_{\text{raw}}) + \sum_{j\neq y}\exp(z_j)}
  = \frac{\sum_{j\neq y}\exp(z_j)}{\exp(r_{\text{raw}}) + \sum_{j\neq y}\exp(z_j)}.
\]
Thus, for \(i \neq y\) we rewrite:
\[
  p_{\text{General CE-U}}(i) = \left(1 - r_{\text{normalized}}\right) \cdot \frac{\exp(z_i)}{\sum_{j\neq y}\exp(z_j)}.
\]
We now define the CE-U probability (which suppresses the true label) for non-true tokens as:
\[
  p_{\text{CE-U}}(i) \coloneqq \frac{\exp(z_i)}{\sum_{j\neq y}\exp(z_j)} \quad \text{for } i \neq y.
\]
Then, for \(i \neq y\) we have:
\[
  p_{\text{General CE-U}}(i) = \left(1 - r_{\text{normalized}}\right) \, p_{\text{CE-U}}(i).
\]

\paragraph{Combining the Two Cases}
To express the entire target distribution compactly, we introduce the one-hot indicator function for the true label:
\[
  \operatorname{one-hot}(y)_i =
  \begin{cases}
    1, & \text{if } i = y, \\
    0, & \text{otherwise}.
  \end{cases}
\]
Thus, the target distribution for any token \(i\) can be written as:
\[
  p_{\text{General CE-U}}(i) = r_{\text{normalized}} \cdot \operatorname{one-hot}(y)_i + \left(1 - r_{\text{normalized}}\right) \cdot p_{\text{CE-U}}(i).
\]

\paragraph{Summary}
In summary, the derivation shows that:
\begin{enumerate}[leftmargin=*]
  \item The normalized score is defined as
    \[
      r_{\text{normalized}} = \frac{\exp(r_{\text{raw}})}{\exp(r_{\text{raw}}) + \sum_{j\neq y}\exp(z_j)},
    \]
  \item The target distribution in the General CE-U framework is an interpolation:
    \[
      p_{\text{General CE-U}}(i) = r_{\text{normalized}} \cdot \operatorname{one-hot}(y)_i + \left(1 - r_{\text{normalized}}\right) \cdot p_{\text{CE-U}}(i),
    \]
    where \(p_{\text{CE-U}}(i)\) is the distribution obtained by suppressing the true label.
\end{enumerate}
This formulation unifies standard cross-entropy (where \(r_{\text{normalized}}=1\)) and the CE-U loss (where \(r_{\text{normalized}}=0\)) by adjusting the coefficient \(r_{\text{normalized}}\) to smoothly interpolate between them.

\subsection{Reference Implementation}
\label{appendix:ceu_code}

\lstset{label=lst:ceu_code}
\lstset{caption={CE-U Loss Function Implementation}}
\begin{python}
from torch import Tensor
import torch.nn.functional as F
def cross_entropy_unlearning_loss(
  logits: Tensor,
  labels: Tensor,
  ignore_index: int = -100,
) -> Tensor:
  """
  Implementation of Cross Entropy Unlearning Loss (CE-U).

  This function creates a modified target distribution by setting the logit corresponding to the true label to negative infinity, effectively forcing the model to assign zero probability to the correct answer. The loss then minimizes the KL divergence between this target distribution and the model's output.

  Args:
    logits: Model output logits with shape [batch_size, sequence_length, vocabulary_size]
    labels: Ground truth token indices with shape [batch_size, sequence_length]
    ignore_index: Token indices to ignore in the loss calculation (typically padding)

  Returns:
    A scalar tensor representing the mean unlearning loss across valid positions
  """
  batch_size, sequence_length, vocabulary_size = logits.shape
  # Extract valid logits and labels based on ignore_index.
  if ignore_index is not None:
  # Shape: [batch_size, sequence_length], boolean mask
  valid_mask = labels != ignore_index
    # Shape: [num_valid_positions, vocabulary_size]
    valid_logits = logits[valid_mask]
    # Shape: [num_valid_positions]
    valid_labels = labels[valid_mask]
  else:
    # Shape: [batch_size*sequence_length, vocabulary_size]
    valid_logits = logits.view(-1, vocabulary_size)
    # Shape: [batch_size*sequence_length]
    valid_labels = labels.view(-1)

  # Create a copy of valid_logits to generate the target distribution
  # Shape: [num_valid_positions, vocabulary_size]
  valid_target_logits = valid_logits.detach().clone()

  # Suppress the logits corresponding to the true token by setting them to -inf.
  # This ensures that the probability for the true token is effectively zero after softmax.
  valid_target_logits.scatter_(
    dim=-1,
    index=valid_labels.unsqueeze(-1),  # Shape: [num_valid_positions, 1]
    value=float("-inf"),
  )  # Result shape: [num_valid_positions, vocabulary_size]

  # Apply softmax to generate the target probability distribution
  # Shape: [num_valid_positions, vocabulary_size]
  valid_target_probabilities = F.softmax(valid_target_logits, dim=-1)

  # Compute the cross entropy loss between input logits and target probabilities
  # The loss is averaged over the valid positions and returns a scalar tensor
  return F.cross_entropy(
    input=valid_logits,
    target=valid_target_probabilities,
  )
\end{python}

\lstset{label=lst:general_ceu_code}
\lstset{caption={General CE-U Loss Function Implementation}}
\begin{python}
from torch import Tensor
import torch.nn.functional as F
def cross_entropy_unified_loss(
  logits: Tensor,
  labels: Tensor,
  scores: Tensor,
  ignore_index: int = -100,
  use_raw_scores: bool = False,
) -> Tensor:
  """
  Implementation of General Cross Entropy Unified Loss (General CE-U).

  This function creates a target distribution that can smoothly transition between:
  - Standard supervised learning (when scores = 1 or raw scores = +inf)
  - Cross entropy unlearning (when scores = 0 or raw scores = -inf)
  - Intermediate reinforcement learning from preferences (when 0 < scores < 1)

  Args:
    logits: Model output logits with shape [batch_size, sequence_length, vocabulary_size]
    labels: Ground truth token indices with shape [batch_size, sequence_length]
    scores: Score values (importance weights) for each valid position with shape [num_valid_positions]
    ignore_index: Token indices to ignore in the loss calculation (typically padding)
    use_raw_scores: If True, scores are treated as direct logit values in log-space; If False, scores are treated as probabilities in [0,1] for interpolation

  Returns:
    A scalar tensor representing the mean unified loss across valid positions
  """
  batch_size, sequence_length, vocabulary_size = logits.shape
  # Extract valid logits and labels based on ignore_index
  if ignore_index is not None:
  # Shape: [batch_size, sequence_length], boolean mask
  valid_mask = labels != ignore_index
  # Shape: [num_valid_positions, vocabulary_size]
  valid_logits = logits[valid_mask]
  # Shape: [num_valid_positions]
  valid_labels = labels[valid_mask]
  else:
  # Shape: [batch_size*sequence_length, vocabulary_size]
  valid_logits = logits.view(-1, vocabulary_size)
  # Shape: [batch_size*sequence_length]
  valid_labels = labels.view(-1)

  if use_raw_scores:
    # Create target logits directly using raw scores
    # Start with a copy of the original logits
    # Shape: [num_valid_positions, vocabulary_size]
    valid_target_logits = valid_logits.detach().clone()

    # Set the logits for true labels directly to the provided raw scores
    # This provides fine-grained control over true label probabilities in the target distribution
    valid_target_logits.scatter_(
      dim=-1,
      index=valid_labels.unsqueeze(-1),  # Shape: [num_valid_positions, 1]
      value=scores.unsqueeze(-1),  # Shape: [num_valid_positions, 1]
    )  # Result shape: [num_valid_positions, vocabulary_size]

    softmax_probabilities = F.softmax(valid_target_logits, dim=-1)
    one_hot_probabilities = F.one_hot(valid_labels, num_classes=vocabulary_size).float()
    mask = torch.isinf(scores)  # Shape: [num_valid_positions]
    valid_target_probabilities = torch.where(
      mask.unsqueeze(-1), one_hot_probabilities, softmax_probabilities
    )
  else:
    # Create an unlearning distribution by suppressing true labels
    # Shape: [num_valid_positions, vocabulary_size]
    valid_unlearning_logits = valid_logits.detach().clone()

    # Set the logits for true labels to -inf to ensure zero probability
    valid_unlearning_logits.scatter_(
      dim=-1,
      index=valid_labels.unsqueeze(-1),  # Shape: [num_valid_positions, 1]
      value=float("-inf"),
    )  # Result shape: [num_valid_positions, vocabulary_size]

    # Compute the unlearning probability distribution
    # Shape: [num_valid_positions, vocabulary_size]
    valid_unlearning_probabilities = F.softmax(valid_unlearning_logits, dim=-1)

    # Create the target distribution as an interpolation between:
    # - The unlearning distribution (when scores = 0)
    # - The one-hot ground truth distribution (when scores = 1)
    # scores.unsqueeze(-1) has shape: [num_valid_positions, 1]
    # F.one_hot(...) has shape: [num_valid_positions, vocabulary_size]
    valid_target_probabilities = (
      valid_unlearning_probabilities * (1 - scores.unsqueeze(-1)) +
      F.one_hot(valid_labels, num_classes=vocabulary_size) *
      scores.unsqueeze(-1)
    )  # Shape: [num_valid_positions, vocabulary_size]

  # Compute the cross entropy loss between input logits and target probabilities
  # The loss is averaged over the valid positions and returns a scalar tensor
  return F.cross_entropy(
    input=valid_logits,
    target=valid_target_probabilities,
  )
\end{python}

\end{document}